\documentclass[10pt,twocolumn,letterpaper]{article}
\pdfoutput=1
\usepackage{cvpr}
\usepackage{times}
\usepackage{epsfig}
\usepackage{graphicx}
\usepackage{amsmath}
\usepackage{amssymb}
\usepackage{xcolor}
\usepackage{mathtools}
\usepackage{amsmath}
\usepackage{amsfonts}
\usepackage[normalem]{ulem}
\useunder{\uline}{\ul}{}
\usepackage{graphicx} % for '\resizebox` macro
\usepackage{tabularx} % for 'tabularx' environment
\usepackage{setspace}
\usepackage{pifont}
\usepackage{multirow}
\usepackage{subfig}
\usepackage{caption}
\usepackage[misc]{ifsym}

\newcommand{\bluetwo}{\textcolor{black}}
\newcommand{\new}{\textcolor{black}}

\newcommand{\red}{\textcolor{black}}
\newcommand{\blue}{\textcolor{black}}
\newcommand{\green}{\textcolor{black}}
\usepackage[breaklinks=true,bookmarks=false]{hyperref}

\cvprfinalcopy % *** Uncomment this line for the final submission

\def\cvprPaperID{9591} % *** Enter the CVPR Paper ID here

\ifcvprfinal\fi
\begin{document}

%%%%%%%%% TITLE
\title{Old is $\mathbf{\mathcal{G}^{old}}$: Redefining the Adversarially 
Learned One-Class Classifier Training Paradigm}

\author{Muhammad Zaigham Zaheer$^{1,2}$, Jin-ha Lee$^{1,2}$, Marcella Astrid$^{1,2}$, Seung-Ik Lee$^{1,2}$\\
$^{1}$University of Science and Technology, $^{2}$Electronics and Telecommunications Research Institute,\\
Daejeon, South Korea\\
{\tt\small \{mzz, jhlee, marcella.astrid\}@ust.ac.kr,}
{\tt\small  the\_silee@etri.re.kr}
%\and
}

\maketitle
\thispagestyle{empty}

%%%%%%%%% ABSTRACT
\begin{abstract}
A popular method for anomaly detection is \bluetwo{to use} \blue{the} generator of an adversarial network to formulate anomaly score over reconstruction loss of input. Due to the rare occurrence of anomalies, optimizing such networks can be a cumbersome task. Another possible approach is to use both generator and discriminator for anomaly detection. However, attributed to the involvement of adversarial training, this model is often unstable in a way that the performance fluctuates drastically with each training step. In this \bluetwo{study}, we propose a framework that effectively generates stable results across a wide range of training steps and allows us to use both \blue{the} generator and the discriminator of an adversarial model for efficient and robust anomaly detection. Our approach transforms the fundamental role of a discriminator from identifying real and fake data to distinguishing between good and bad quality reconstructions.
% The training examples are created using only normal training data by employing current generator as well as an old state of the same generator. 
\blue{To this end, we prepare training examples for the good quality reconstruction by employing the current generator, whereas poor quality examples are obtained by utilizing an old state of the same generator.}
This way, the discriminator learns to detect subtle distortions that often appear in reconstructions of the anomaly inputs. Extensive experiments performed on Caltech-256 and MNIST image datasets for novelty detection show superior results. Furthermore, on UCSD Ped2 video dataset for anomaly detection, our model achieves a frame-level AUC of 98.1\%, surpassing recent state-of-the-art methods.
\end{abstract}

%%%%%%%%% BODY TEXT
\section{Introduction}
\label{section:introduction}
Due to rare occurrence of anomalous scenes, the anomaly detection problem is usually seen as one-class classification (OCC) in which only normal data is used to learn a novelty detection model \cite{liu2018future_novelty,zhang2016video_novelty,luo2017revisit_novelty,xia2015learning_novelty_fig5,zaheer2018ensemble,hinami2017joint_novelty,sultani2018real_novelty,sabokrou2017deep_novelty,hasan2016learning_novelty,smeureanu2017deep_novelty,ravanbakhsh2018plug_novelty,ravanbakhsh2017abnormal_novelty}. One of the recent trends to learn one-class data is by using an encoder-decoder architecture such as denoising auto-encoder \cite{sabokrou2018adversarially_alocc,xu2015learning_denoise,xu2017detecting_denoise,vincent2008extracting_denoise}. Generally, in this scheme, training is carried out until the model starts to produce good quality reconstructions \cite{sabokrou2018adversarially_alocc,Shama_2019_ICCV_good}. \bluetwo{During the test time,} it is expected to show high reconstruction loss for abnormal data which corresponds to a \blue{high} anomaly score. With the recent developments in Generative Adversarial Networks (GANs) \cite{goodfellow2014generative_gan}, \blue{some} researchers \blue{also} explored the possibility of improving the generative results using adversarial training \cite{Shama_2019_ICCV_good,radford2015unsupervised_good_adversarial}. Such training fashion substantially enhances the data regeneration quality \cite{pathak2016context_adversarial_good,goodfellow2014generative_gan,Shama_2019_ICCV_good}. \bluetwo{At the test time,} the trained generator $\mathcal{G}$ is then decoupled from  the discriminator $\mathcal{D}$ to be used as a reconstruction model.
% in a similar fashion as in any conventional encoder-decoder based anomaly detection method.
As reported in \cite{sabokrou2018adversarially_alocc,ravanbakhsh2017abnormal_novelty,ravanbakhsh2019training}, a wide difference in reconstruction loss between the normal and abnormal data can be achieved due to adversarial training, \bluetwo{which results} in a better anomaly detection system. However, relying only on the reconstruction capability of a generator does not \new{oftentimes work well because the usual encoder-decoder style generators may unexpectedly well-reconstruct the unseen data which drastically degrades the anomaly detection performance.}

\begin{figure}[t]
\begin{center}
   \includegraphics[width=0.8\linewidth]{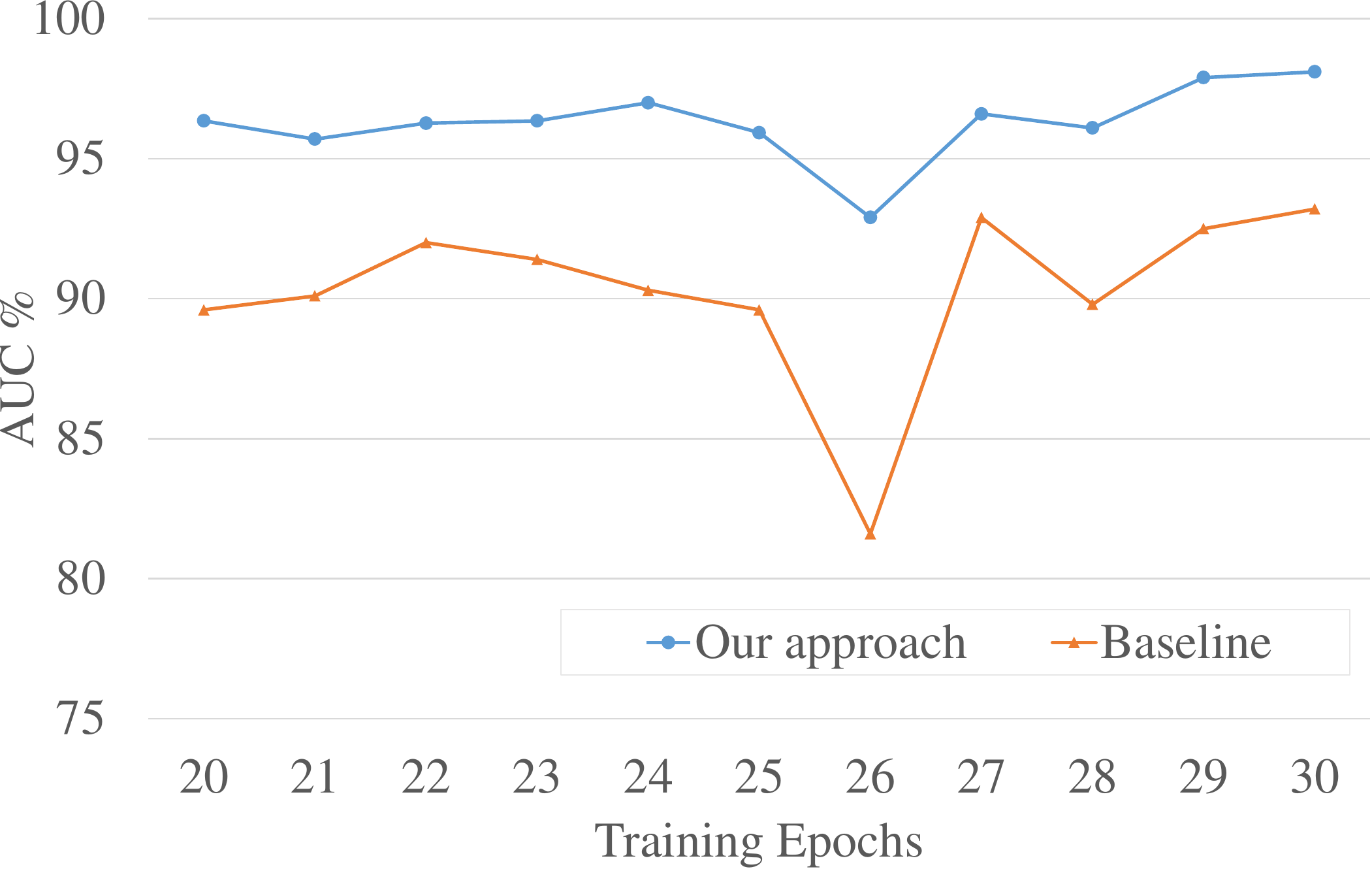}
\end{center}
   \caption{\blue{Dynamics of AUC performance over training epochs: The baseline shows high fluctuations while our approach not only shows stability across various epochs but also yields higher AUC.}}
\label{fig:AUC_Comparison}
\end{figure}

A natural drift in this domain is towards the idea of using $\mathcal{D}$ along with the conventional utilization of $\mathcal{G}$ for anomaly detection. The intuition is to gain maximum benefits of the one-class adversarial training by utilizing both $\mathcal{G}$ and $\mathcal{D}$ instead of only $\mathcal{G}$. However, this also brings along the problems commonly associated with such architectures.  \red{For example}, defining a criteria to stop the training is still a challenging problem \cite{goodfellow2014generative_gan,pathak2016context_adversarial_good}.
As discussed in Sabokrou \etal. \cite{sabokrou2018adversarially_alocc}, the performance of such adversarially learnt one-class classification architecture is highly dependent on the criteria of when to halt the training. 
% Undesirable results will be generated in the case of stopping it prematurely (undertrained $\mathcal{G}$) or in the case of overtraining (confused $\mathcal{D}$ because of the real-looking fake data, \red{including anomalies}, being generated by $\mathcal{G}$) \red{\cite{sabokrou2018adversarially_alocc}}. 
\blue{In the case of stopping prematurely, $\mathcal{G}$ will be undertrained and in the case of overtraining, $\mathcal{D}$ \bluetwo{may get confused} because of the real-looking fake data.}
Our experiments show that a $\mathcal{G}$+$\mathcal{D}$ trained as a collective model for anomaly detection \blue{(referred to as a baseline)} will not ensure higher convergence at any arbitrary training step over its predecessor. \red{Figure~\ref{fig:AUC_Comparison}} shows frame-level area under the curve (AUC) performance of the baseline over several epochs of training on UCSD Ped2 dataset \red{\cite{chan2008ucsd}}.
%This architecture is a conventional denoising encoder-decoder\red{\cite{sabokrou2018adversarially_alocc, xu2015learning_denoise,xu2017detecting_denoise,vincent2008extracting_denoise}} trained with an adversarial discriminator.
\blue{Although we get high performance peaks at times, it can be seen that the performance fluctuates substantially even between two arbitrary consecutive epochs.} 
% It is nevertheless interesting to notice that at various training stages, performance of the network is considerably good (\red{AUC $>$ 92 \%}) however, given that a novelty detection model is trained only on normal data and has no access to abnormal data for training or validation whatsoever, it becomes cumbersome to optimize the training of an adversarial network for such problems. 
Based on these findings, 
% as well as the discussion provided in \red{\cite{sabokrou2018adversarially_alocc}},
 it can be argued that a $\mathcal{D}$ as we know it, may not be a suitable choice in a one-class classification problem, such as anomaly detection.

%====================================== AUC comparison diagram
%------------------------------------------------------------------------
\blue{Following this intuition, }we devise an approach \blue{for training of an adversarial network towards anomaly detection by transforming the basic role of $\mathcal{D}$} from distinguishing between real and fake to identifying good and bad quality reconstructions. This property of $\mathcal{D}$ is highly desirable \blue{in anomaly detection because} a trained $\mathcal{G}$ \blue{would} not produce as good reconstruction for \blue{abnormal} data as it would for the normal data conforming to the learned representations. To this end we propose a two-stage training process. Phase one is identical to the common practice of training an adversarial denoising auto-encoder \cite{pathak2016context_adversarial_good,xu2017detecting_denoise,vincent2008extracting_denoise}. Once $\mathcal{G}$ achieves a reasonably trained state (i.e. showing low reconstruction losses), we begin \green{phase two} in which $\mathcal{D}$ is optimized by training on various good quality and bad quality reconstruction examples. \blue{Good quality reconstruction examples come from real data as well as the data regenerated by $\mathcal{G}$, whereas bad reconstruction examples are obtained by utilizing an old state of the generator ($\mathcal{G}^{old}$) as well as by using our proposed pseudo-anomaly module. \bluetwo{Shown in Figure~\ref{fig:pseudo_reconstruction_module}, this pseudo-anomaly module makes use of the training data to create} \textit{anomaly-like} examples}. With this two-phase training process, we expect $\mathcal{D}$ to be trained in such a way that it can robustly discriminate reconstructions coming from normal and abnormal data. As shown in Figure~\ref{fig:AUC_Comparison}, our model not only provides superior performance but also shows stability across several training epochs.

In summary, the contributions of our paper are as follows: 1) \bluetwo{this} work is among the first few to employ $\mathcal{D}$ along with $\mathcal{G}$ at test time for anomaly detection. Moreover, to the best of our knowledge, it is the first one to extensively report the impacts of \new{using the conventional $\mathcal{G} + \mathcal{D}$ formulation and the consequent instability}. 2) Our approach of transforming the role of a discriminator towards anomaly detection problem by utilizing \red{an} old state \blue{$\mathcal{G}^{old}$} of the generator along with the proposed pseudo-anomaly module, substantially improves stability of the system. Detailed \bluetwo{analysis} provided in this paper show\bluetwo{s} that our model is independent of a hard stopping criteria and achieves consistent results over a wide range of training epochs.
3) Our method outperforms state-of-the-art \cite{sabokrou2018adversarially_alocc, ionescu2019object, ren2015unsupervised,Nguyen_2019_ICCV, Gong_2019_ICCV, tudor2017unmasking_novelty,luo2017revisit_novelty,nguyen2019hybrid,hinami2017joint_novelty,liu2018classifier_novelty,ravanbakhsh2017abnormal_novelty,luo2017remembering,ravanbakhsh2018plug_novelty,sun2017online,hasan2016learning_novelty,liu2018future_novelty,xu2015learning_denoise,zhang2016video_novelty, zhao2017spatio} in the experiments conducted \new{on} MNIST \cite{mnist} and Caltech-256 \cite{griffin2007caltech} datasets for novelty detection as well as \new{on} UCSD Ped2 \cite{chan2008ucsd} video dataset for anomaly detection.
Moreover, on the latter dataset, our approach provides a substantial absolute gain of 5.2\% over the baseline method achieving frame level AUC of 98.1\%.
%===========================ARCHITECTURE_DIAGRAM
\begin{figure*}
\begin{center}
  \includegraphics[width=0.97\linewidth]{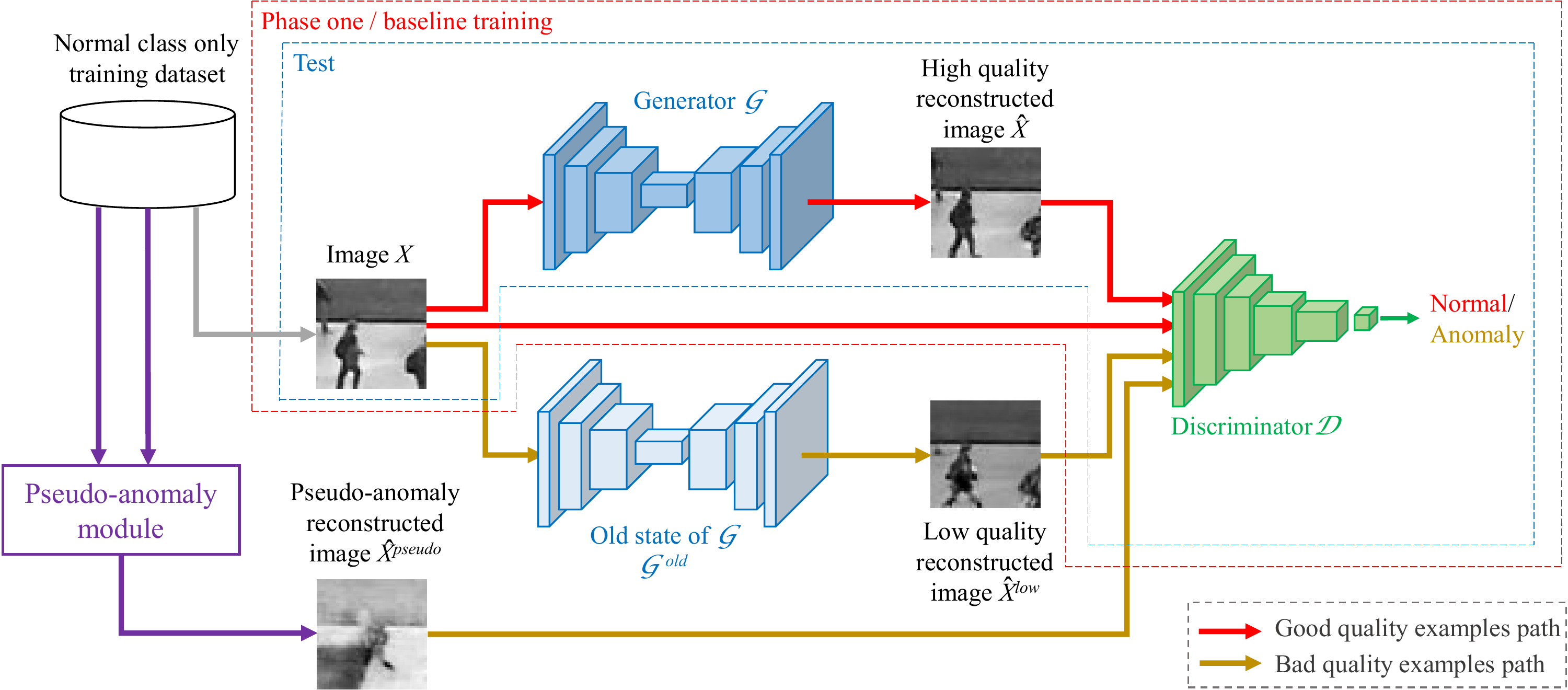}
   \caption{Our proposed \blue{OGNet} framework. Phase one is the baseline training, carried out to obtain a reasonably trained state of $\mathcal{G}$ and $\mathcal{D}$. A frozen low epoch state ($\mathcal{G}^{old}$) of the generator is stored during this training. In phase two, only $\mathcal{D}$ is updated to distinguish between good and bad quality reconstructions. \new{Good quality examples correspond to real training images as well as the images reconstructed using $\mathcal{G}$ while bad quality examples are obtained using $\mathcal{G}^{old}$ as well as the proposed pseudo-anomaly module. This module assists $\mathcal{D}$ to learn the underlying patterns of anomalous input reconstructions. During test, inferences are carried out through $\mathcal{G}$ and $\mathcal{D}$ only and the output of $\mathcal{D}$ is considered as anomaly score. Best viewed in color.}}
\label{fig:architecture}
\end{center}
\vspace{-4mm}
\end{figure*}

\section{Related Work}
\blue{Anomaly detection is often seen as a novelty detection problem \cite{liu2018future_novelty,zhang2016video_novelty,luo2017revisit_novelty,hinami2017joint_novelty,xia2015learning_novelty_fig5,sultani2018real_novelty,sabokrou2017deep_novelty,bergadano2019keyed,hasan2016learning_novelty,smeureanu2017deep_novelty,ravanbakhsh2018plug_novelty,ravanbakhsh2018plug_novelty,ravanbakhsh2017abnormal_novelty} in which a model is \bluetwo{trained} based on the known normal class to ultimately detect unknown outliers as abnormal. To simplify the task, some works proposed to use object tracking \cite{wang2014learning_realworld38,basharat2008learning_realworld7,medioni2001event_twostream34,piciarelli2008trajectory_twostream36,zhang2009learning_twostream53} or motion \cite{kratz2009anomaly_realworld26,hou2017tube_realworld20,cui2011abnormal_realworld10}. Handpicking features in such a way can often deteriorate the performance significantly. With the increased popularity of deep learning, some researchers \cite{smeureanu2017deep_novelty,ravanbakhsh2017abnormal_novelty} also proposed to use pre-trained convolution network based features to train one-class classifiers. Success of such methods is highly dependent on the base model which is often trained on some unrelated datasets.}

\blue{A relatively new addition to the field, image regeneration based works \cite{Gong_2019_ICCV,ren2015unsupervised,xu2015learning_denoise,ionescu2019object,Nguyen_2019_ICCV,nguyen2019hybrid,xu2017detecting_denoise,sabokrou2017deep_novelty} are the ones that make use of a generative network to learn features in an unsupervised way. Ionescu \etal in \cite{ionescu2019object} proposed to use convolutional auto-encoders on top of object detection to learn motion and appearance representations. Xu \etal. \cite{xu2015learning_denoise,xu2017detecting_denoise} used a one-class SVM learned using features from stacked auto-encoders. \green{Ravanbakhsh \etal\cite{ravanbakhsh2017abnormal_novelty}} used generator as a reconstructor to detect abnormal events \bluetwo{assuming that a generator is unable to} reconstruct the inputs that do not conform the normal training data. \green{In \cite{Nguyen_2019_ICCV,nguyen2019hybrid},  the authors} suggested to use a cascaded decoder to learn motion as well as appearance from normal videos. \bluetwo{However, in all these schemes, only a generator is employed to perform detection.} \red{Pathak \etal \cite{pathak2016context_adversarial_good}} \bluetwo{proposed} adversarial training to enhance the quality of regeneration. However, \bluetwo{they also discard the discriminator} once the training is finished.} \blue{A unified generator and discriminator model for anomaly detection is proposed in Sabokrou \etal\cite{sabokrou2018adversarially_alocc}. The model shows promising results, however it is \bluetwo{often} not stable and the performance relies heavily on the criteria to stop training. \bluetwo{Recently, Shama \etal\cite{Shama_2019_ICCV_good} proposed an idea of utilizing output of an adversarial discriminator to increase the image quality of generated images. Although not related to anomaly detection, it provides an interesting intuition to make use of both adversarial components for an enhanced performance.}}

\bluetwo{Our work, although built on top of an unsupervised generative network,} is different from the approaches in \cite{Gong_2019_ICCV,ionescu2019object,Nguyen_2019_ICCV,nguyen2019hybrid,xu2017detecting_denoise,sabokrou2017deep_novelty} as we explore to utilize the unified generator and discriminator model for anomaly detection. The most similar work to ours is by Sabokrou \etal\cite{sabokrou2018adversarially_alocc} \bluetwo{and Lee \etal \cite{lee2018stan} as they also explore the possibility of using discriminator, along with the conventional usage of generator, for anomaly detection. However, our approach is substantially different from these. In \cite{sabokrou2018adversarially_alocc}, a conventional adversarial network is trained based on a criteria to stop the training whereas, in \cite{lee2018stan}, an LSTM based approach is utilized for training. In contrast,} we utilize a pseudo-anomaly module along with an old state of the generator, to modify the ultimate role of a discriminator from distinguishing between real and fake to detecting between and bad quality reconstructions. This way, our overall framework, although trained adversarially in the beginning, finally aligns both the generator and the discriminator to complement each other towards anomaly detection.
\section{Method} \label{section:method}
In this section, we present our \red{OGNet} framework. As described in \bluetwo{Section \ref{section:introduction}}, most of the existing \new{GANs based anomaly} detection approaches \new{completely discard discriminator at test time and use generator only. Furthermore, even if both models are used,} the unavailability of a criteria to stop the training coupled with the instability \blue{over training epochs} caused by adversary makes the convergence uncertain. We \new{aim} to change that by redefining the role of a discriminator to make it more suitable for anomaly detection problems. Our solution is generic, hence it can be integrated with any existing one-class adversarial networks.
\subsection{Architecture Overview}
In order to maintain consistency and to have a fair comparison, we kept our baseline architecture similar to the one proposed by \green{Sabokrou \etal\cite{sabokrou2018adversarially_alocc}}. The generator $\mathcal{G}$, a typical denoising auto-encoder, is coupled with the discriminator $\mathcal{D}$ to learn one class data in an unsupervised adversarial fashion. The goal of this model is to play a min-max game to optimize the following objective function:

\begin{equation}
\begin{multlined}
\underset{\mathcal{G}}{\text{min}} \: \underset{\mathcal{D}}{\text{max}} \:  \Bigl(\mathbb{E}_{X \sim p_t}[\text{log}(1-\mathcal{D}(X))] \\ + \mathbb{E}_{\tilde{X} \sim p_t + \mathcal{N}_\sigma}[\text{log}(\mathcal{D}(\mathcal{G}(\tilde{X})))]\Bigr),
\end{multlined}
\label{eq:jointGAN}
\end{equation}
\red{where ${\tilde{X}}$ is the input image} $X$ with added noise \new{$\mathcal{N}_\sigma$} as in a typical denoising auto-encoder. Our model, built on top of the baseline, makes use of an old frozen generator (\red{$\mathcal{G}^{old}$}) to create low quality reconstruction examples. We also propose a pseudo-anomaly module to assist $\mathcal{D}$ in learning the behavior of $\mathcal{G}$ in the case of unusual or anomalous input, \blue{which we found out to be very useful towards the robustness of \bluetwo{our approach} (Table \ref{tab:ablation})}. The overall purpose of our proposed framework is to alter the learning paradigm of $\mathcal{D}$ from distinguishing between real and fake to differentiating between good and bad reconstructions. This way, the discriminator gets aligned with the conventional philosophy of generative one-class learning models in which the reconstruction quality of the data from a known class is better than the data from unknown or anomaly classes.
%====================================== pseudo generator
\begin{figure}[t]
\begin{center}
%\fbox{\rule{0pt}{2in} \rule{0.8\linewidth}{0pt}}
   \includegraphics[width=0.80\linewidth]{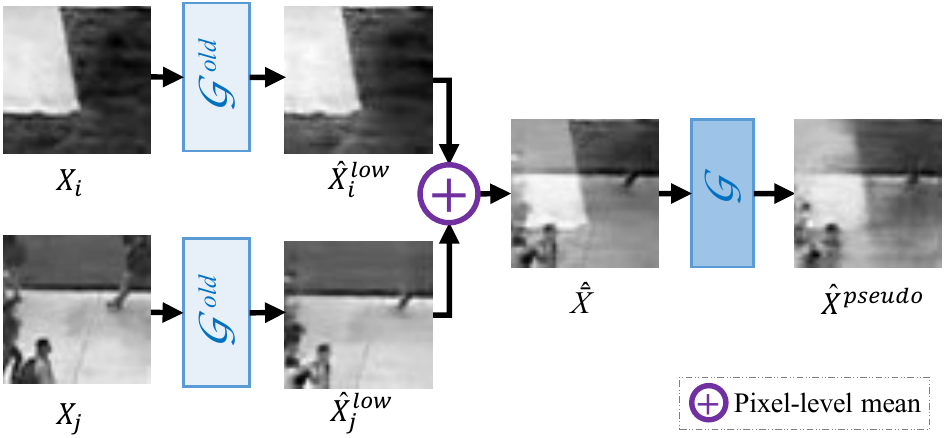}
\end{center}
   \caption{Our proposed pseudo-anomaly module. A pseudo-anomaly $\hat{\bar{X}}$ is created by regenerating two arbitrary training images through $\mathcal{G}^{old}$ followed by a pixel-level mean. \new{Finally, $\hat{X}^{pseudo}$ is created as $\mathcal{G}(\hat{\bar{X}})$ to mimic the regeneration behavior of $\mathcal{G}$ for anomalous inputs.}}
   
%   This helps the $\mathcal{D}$ to learn \green{efficiently(is it efficient? or robust?)} and to eventually complement \green{$\mathcal{G}$(or is it D?)} during the test time for a robust anomaly detection system}}
\label{fig:pseudo_reconstruction_module}
\end{figure}
\subsection{Training}
The training of our model is carried out in two phases (\red{see Figure~\ref{fig:architecture}}). Phase one is similar to the common practices in training an adversarial one-class classifier \cite{sabokrou2018adversarially_alocc,lawson2017finding,schlegl2017unsupervised,ravanbakhsh2019training}. $\mathcal{G}$ tries to regenerate real-looking fake data which is then fed into $\mathcal{D}$ along with real data. The $\mathcal{D}$ learns to discriminate between real and fake data, success or failure of which then becomes a supervision signal for $\mathcal{G}$. This training is carried out until \new{$\mathcal{G}$} starts to create real looking images with a reasonably low reconstruction loss. Overall, phase one minimizes the following loss function:
\begin{equation}
\mathcal{L} = \mathcal{L}_{\mathcal{G}+\mathcal{D}} + \lambda\mathcal{L}_R,
\end{equation}
% \begin{equation}
% \begin{multlined}
% \underset{\mathcal{G}}{\text{min}} \: \underset{\mathcal{D}}{\text{max}} \: \Bigl( \mathbb{E}_{X \sim p_t}[\text{log}(1-\mathcal{D}(X))] \\ + \mathbb{E}_{\tilde{X} \sim p_t + \mathcal{N}_\sigma}\text{log}[(\mathcal{D}(\mathcal{G}(\tilde{X})))] 
% \Bigr) + \lambda_1\mathcal{L}_R
% \end{multlined}
% \end{equation}
where $\mathcal{L}_{\mathcal{G}+\mathcal{D}}$ is the loss function of our joint training objective defined in Equation \ref{eq:jointGAN}, $\mathcal{L}_R = || X - \mathcal{G}(\tilde{X}) ||^2$ is the reconstruction loss, and $\lambda$ is a weighing hyperparameter.
Additionally, as \green{phase one} progresses, we \blue{save} a low-epoch generator model (\red{$\mathcal{G}^{old}$}) for later use in phase two of the training. Deciding which low epoch to be used can be an intuitive selection based on the quality of regeneration. Obviously, we want \red{$\mathcal{G}^{old}$} to generate low quality images compared to \new{a} trained $\mathcal{G}$. \new{However, it is not necessary to select any specific epoch number for this generator. We will prove this empirically in Section \ref{section:experiments} by showing that the final }convergence of our model is not dependent on a strict selection of the epoch numbers and that various generic settings are possible to obtain a $\mathcal{G}^{old}$.

Phase two of the training is where we make use of \bluetwo{the frozen models \red{$\mathcal{G}^{old}$} and $\mathcal{G}$ to update $\mathcal{D}$.}  This way $\mathcal{D}$ starts learning to discriminate between good and bad quality reconstructions, hence becoming suitable for one-class classification problems such as anomaly detection.

Details of the phase two training are discussed next:

\noindent\textbf{Goal.} The essence of phase two training is to provide examples of good quality and bad quality reconstructions to $\mathcal{D}$, with a purpose of making it learn about the kind of output that $\mathcal{G}$ would produce in the case of an unusual input. The training is performed for just a few iterations since the already trained $\mathcal{D}$ converges quickly. A detailed study on this is added in Section \ref{section:experiments}. 

\noindent\textbf{Good quality examples.} $\mathcal{D}$ is provided with real data ($X$), which is the best possible case of reconstruction, and the actual high quality reconstructed data ($\hat{X} = \mathcal{G}(X)$) produced by the trained $\mathcal{G}$ as an example of good quality \bluetwo{examples.}

\noindent\textbf{Bad quality examples.} Examples of low quality reconstruction ($\hat{X}^{low}$) are generated using \red{$\mathcal{G}^{old}$}. In addition, a pseudo-anomaly module, shown in \red{Figure~\ref{fig:pseudo_reconstruction_module}}, is formulated with a combination of \red{$\mathcal{G}^{old}$} and the trained $\mathcal{G}$, which simulates examples of reconstructed pseudo-anomalies ($\hat{X}^{pseudo}$).

\begin{figure}[b]
\begin{center}
\includegraphics[width=0.95\linewidth]{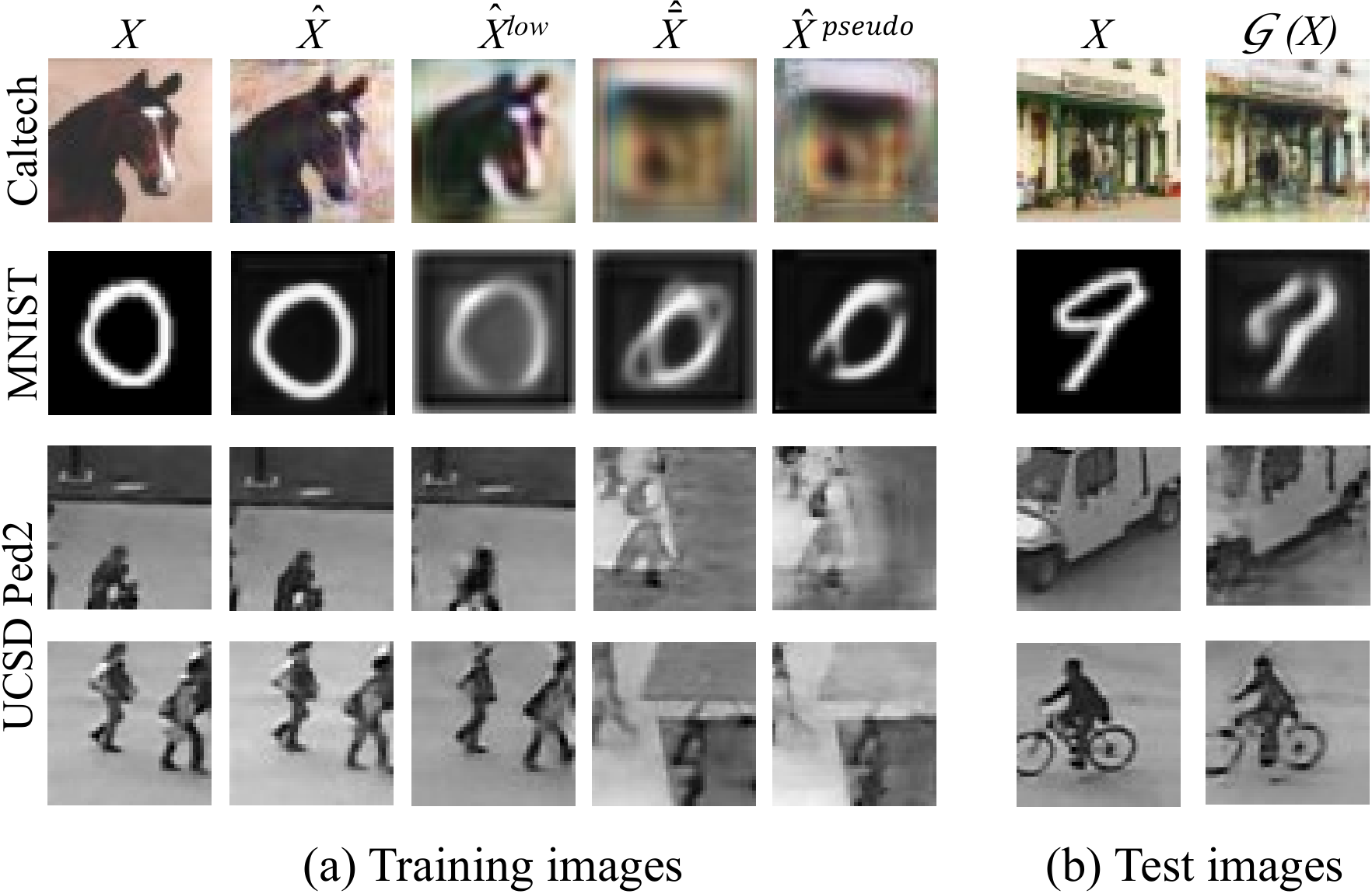}
\end{center}
\caption{\red{Example images from different stages of our framework. (a) Left to right: Original image ($X$), high quality reconstructed ($\hat{X}$), low quality reconstructed ($\hat{X}^{low}$), pseudo anomaly ($\hat{\bar{X}} $), pseudo anomaly reconstructed ($\hat{X}^{pseudo}$). (b) Left column shows outlier / anomaly examples whereas right column shows respective regenerated outputs $\mathcal{G}$($X$)}.}
\label{fig:example_images}
\end{figure}

\noindent\textbf{Pseudo anomaly creation.} Given two arbitrary images $X_i$ and $X_j$ from the training dataset, a pseudo anomaly image $\hat{\bar{X}}$ is generated as:

\begin{equation}\label{eq:fake_anomaly}
\hat{\bar{X}} = \frac{\mathcal{G}^{old}(X_i) + \mathcal{G}^{old}(X_j)}{2}  = \frac{\hat{X}^{low}_i + \hat{X}^{low}_j}{2}, \\
\text{where $i \neq j$.}
\end{equation}
This way, the resultant image can contain diverse variations such as shadows and unusual shapes, which are completely unknown to both $\mathcal{G}$ and $\mathcal{D}$ models.
Finally, as the last step in our pseudo-anomaly module, in order to mimic the behavior of $\mathcal{G}$ \blue{when it gets unusual data as input}, $\hat{\bar{X}}$ is then reconstructed using $\mathcal{G}$ to obtain $\hat{X}^{pseudo}$:

\begin{equation}\label{eq:pseudo_reconstruction}
\hat{X}^{pseudo} = \mathcal{G}(\hat{\bar{X}}).
\end{equation}
Example images at each intermediate step can be seen in \red{Figures~\ref{fig:pseudo_reconstruction_module} and~\ref{fig:example_images}}. 

\noindent\textbf{Tweaking the objective function.} The model in phase two of the training takes the form: 
\begin{equation}
\begin{multlined}
\underset{\mathcal{D}}{\text{max}}\:\Bigl(\alpha\mathbb{E}_{X}[\text{log}(1-\mathcal{D}(X))] +\\(1-\alpha)\mathbb{E}_{\hat{X}}[\text{log}(1-\mathcal{D}(\hat{X}))] + \beta\mathbb{E}_{\hat{X}^{low}}[\text{log}(\mathcal{D}(\hat{X}^{low}))] +\\(1-\beta)\mathbb{E}_{\hat{X}^{pseudo}}[\text{log}(\mathcal{D}(\hat{X}^{pseudo}))]\Bigr),
\end{multlined}
\end{equation}
where $\alpha$ and $\beta$ are the trade-off hyperparameters.

Quasi ground truth for the discriminator in \green{phase one} training is defined as:
\begin{equation}
    GT_{\green{phase\_one}}= 
\begin{cases}
    0      & \text{if input is $X$,}\\
    1      & \text{if input is $\hat{X}$}.
\end{cases}
\end{equation}
However, for \green{phase two} training, it takes the form: 
\begin{equation}
    GT_{\green{phase\_two}}= 
\begin{cases}
    0      & \text{if input is $X$ or $\hat{X}$,}\\
    1      & \text{if input is $\hat{X}^{low}$ or $\hat{X}^{pseudo}$}.
\end{cases}
\end{equation}
%\noindent\textbf{Complexity of the learning parameters.} There is no additional change in the number of learning parameters as our baseline model remains same throughout the process.
\subsection{Testing}
At test time, as shown in \red{Figure~\ref{fig:architecture}}, only $\mathcal{G}$ and $\mathcal{D}$ are utilized for one-class classification (OCC). Final classification decision for an input \new{image $X$} is given as:
\begin{equation}
OCC = 
\begin{cases}
    \text{normal class}      & \text{if $\mathcal{D}(\mathcal{G}(X)) < \tau$,}\\
    \text{anomaly class}      & \text{otherwise.}
\end{cases}
\end{equation}
where $\tau$ is a predefined threshold. %At the test time, we do not add noise to the image.
%-----------------------------------------------------------------------------
%====================================== Training and Testing Examples

\section{Experiments} \label{section:experiments}
The evaluation of our \blue{OGNet framework} on three different datasets is reported in this section. \new{Detailed analysis of the performance and its comparison with the state-of-the-art methodologies is also reported. In addition, we provide \bluetwo{extensive} discussion and ablation studies to show the stability as well as the significance of our proposed scheme.} In order to keep the experimental setup consistent with the existing works \cite{liu2018future_novelty,zhang2016video_novelty,luo2017revisit_novelty,xia2015learning_novelty_fig5,hinami2017joint_novelty,sultani2018real_novelty,sabokrou2017deep_novelty,hasan2016learning_novelty,smeureanu2017deep_novelty,ravanbakhsh2018plug_novelty,ravanbakhsh2017abnormal_novelty,sabokrou2018adversarially_alocc,ionescu2019object,Gong_2019_ICCV,nguyen2019hybrid,Nguyen_2019_ICCV}, we tested our method for the detection of outlier images as well as video anomalies.

\noindent\textbf{Evaluation criteria.} Most of our results are formulated based on area under the curve (AUC) computed at frame level due to its popularity in related works \cite{tudor2017unmasking_novelty,luo2017revisit_novelty,nguyen2019hybrid,hinami2017joint_novelty,liu2018classifier_novelty,ravanbakhsh2017abnormal_novelty,luo2017remembering,Gong_2019_ICCV,ravanbakhsh2018plug_novelty,sun2017online,hasan2016learning_novelty,liu2018future_novelty,xu2015learning_denoise,Nguyen_2019_ICCV,zhang2016video_novelty,ionescu2019object,zhao2017spatio,sabokrou2018adversarially_alocc}. Nevertheless, following the evaluation methods adopted in \cite{tsakiris2015dual,lerman2015robust,xu2010robust,rahmani2017coherence,liu2010robust,you2017provable_novelty,sabokrou2018adversarially_alocc,sabokrou2016video,ravanbakhsh2017abnormal_novelty,ravanbakhsh2019training,xu2015learning_denoise,sabokrou2015real,sabokrou2017deep_novelty} we also report $F_1$ score and Equal Error Rate (EER) of our approach.

\noindent\textbf{Parameters and implementation details.} \blue{The implementation is done in  PyTorch} \cite{paszke2017automatic} and the source code is provided at \emph{https://github.com/xaggi/OGNet}.
Phase one of the training in our reports is performed \new{from} 20 to 30 epochs. \new{These numbers are} chosen because the baseline shows high performance peaks within this range (Figure~\ref{fig:AUC_Comparison}). We train on Adam \cite{kingma2014adam} with the learning rate of generator and discriminator in all these epochs set to $10^{-3}$ and $10^{-4}$, respectively. \green{Phase two} of the training is done for $75$ iterations with the learning rate of the discriminator reduced to half. $\lambda, \alpha, \text{ and }\beta$ are set to 0.2, 0.1, 0.001, respectively. Until stated otherwise, default settings of our experiments are set to the aforementioned values. However, for the detailed evaluation provided in a later part of this section, we also conducted experiments and reported results on a range of epochs and iterations for both phases of the training, respectively. Furthermore, \new{until specified otherwise,} we pick the generator after $1^{st}$ epoch and freeze it as \red{$\mathcal{G}^{old}$}. This selection is arbitrary and \new{solely} based on the intuition explained in Section \ref{section:method}. Additionally, in a later part of this section, we also present a robust and generic method to formulate \red{$\mathcal{G}^{old}$} without \new{any need of} handpicking an epoch.
\begin{table*}[]
\begin{center}
\resizebox{\linewidth}{!}{%
\begin{tabular}{ccccccccc}
                     & DPCP\cite{tsakiris2015dual}                                             & REAPER\cite{lerman2015robust} & OutlierPersuit\cite{xu2010robust} & CoP\cite{rahmani2017coherence}   & LRR\cite{liu2010robust}   & R-graph\cite{you2017provable_novelty}     & ALOCC\cite{sabokrou2018adversarially_alocc}       & Ours           \\ \hline
\multicolumn{1}{c|}{AUC} & \begin{tabular}[c]{@{}c@{}}78.3\%\end{tabular} & 81.6\%  & 83.7\%          & 90.5\% & 90.7\% & {\ul 94.8\%} & 94.2\%       & \textbf{98.2\%} \\
\multicolumn{1}{c|}{$F_1$}  & 78.5\%                                            & 80.8\%  & 82.3\%          & 88.0\%  & 89.3\% & 91.4\%       & {\ul 92.8\%} & \textbf{95.1\%} \\ \hline
\multicolumn{1}{c|}{AUC} & 79.8\%                                            & 79.6\%  & 78.8\%          & 67.6\% & 47.9\% & 92.9\%       & {\ul 93.8\%} & \textbf{97.7\%} \\
\multicolumn{1}{c|}{$F_1$}  & 77.7\%                                            & 78.4\%  & 77.9\%          & 71.8\% & 67.1\% & 88.0\%        & {\ul 91.3\%} & \textbf{91.5\%} \\ \hline
\multicolumn{1}{c|}{AUC} & 67.6\%                                            & 65.7\%  & 62.9\%          & 48.7\% & 33.7\% & 91.3\%       & {\ul 92.3\%} & \textbf{98.1\%} \\
\multicolumn{1}{c|}{$F_1$}  & 71.5\%                                            & 71.6\%  & 71.1\%          & 67.2\% & 66.7\% & 85.8\%       & {\ul 90.5\%} & \textbf{92.8\%} \\ \hline
\end{tabular}}
\end{center}
\caption{\red{AUC and $F_1$ score performance  comparison of our framework on Caltech-256 \cite{griffin2007caltech} with the other state of the art methods. Following the existing work \cite{you2017provable_novelty}, each subgroup of rows from top to bottom shows evaluation scores on inliers coming from 1, 3, and 5 \green{different random} classes respectively (best performance as bold and second best as underlined).}}
\label{tab:auc_f1_caltech}
\end{table*}
\subsection{Datasets}
\noindent\textbf{Caltech-256.} This dataset \cite{griffin2007caltech} contains a total of 30,607 images belong to 256 object classes and one `clutter' class. Each category has different number of images, as low as 80 and as high as 827. \new{In order} to perform our experiments, we used the same setup as described in previous works \cite{tsakiris2015dual,lerman2015robust,xu2010robust,rahmani2017coherence,liu2010robust,you2017provable_novelty,sabokrou2018adversarially_alocc}. In a series of three experiments, at most 150 images \new{belong to} 1, 3, and 5 randomly chosen classes are defined as training (inlier) data. Outlier images for test are taken from the ‘clutter’ class in such a way that each experiment has exactly 50\% ratio of \blue{outliers and inliers.}

% it is dataset of..... 0 to 9 \cite{}.
\noindent\textbf{MNIST.} \green{This dataset \cite{mnist} consists of} 60,000 handwritten digits from 0 to 9. The setup to evaluate our method on this dataset is also kept consistent with the previous works \cite{xia2015learning_novelty_fig5,breunig2000lof_fig5,sabokrou2018adversarially_alocc}. In a series of experiments, each category of digits is individually taken as inliers. Whereas, randomly sampled images of the other categories with a proportion of 10\% to 50\% are taken as outliers.

\noindent\textbf{USCD Ped2.} This dataset \cite{chan2008ucsd} comprises of 2,550 frames in 16 training and 2,010 frames in 12 test videos. Each frame is of $240 \times 360$ pixels resolution. \new{Pedestrians dominate most of the frames whereas anomalies include skateboards, vehicles, bicycles, etc}. Similar to \cite{tudor2017unmasking_novelty,luo2017revisit_novelty,nguyen2019hybrid,hinami2017joint_novelty,liu2018classifier_novelty,ravanbakhsh2017abnormal_novelty,luo2017remembering,Gong_2019_ICCV,ravanbakhsh2018plug_novelty,sun2017online,hasan2016learning_novelty,liu2018future_novelty,xu2015learning_denoise,Nguyen_2019_ICCV,zhang2016video_novelty,ionescu2019object,zhao2017spatio,sabokrou2018adversarially_alocc}, frame-level AUC and EER metrics are adopted to evaluate \new{performance} on this dataset.
%An older version of this dataset, ped1 [\red{reference}] is also available. However, the resolution of this version, $158 \times 238$ pixels, is significantly lower. Also, as discussed in [\red{Object centric paper and the reference mentioned in object centric paper}], the results reported in the existing works are not consistent. Some works reported evaluation only on a subset of the test videos [\red{references from object centric paper}] while others reported on full dataset [\red{references from object centric paper}]. Hence, we consider only Ped2 for the evaluation of our approach.
\subsection{Outlier Detection in Images}
\blue{One of the significant applications of a one-class learning algorithm is outlier detection.} In this problem, objects belonging to known classes are treated as inliers based on which the model is trained. Other objects that do not belong to these classes are treated as outliers, which the model is supposed to detect based on its training. 
Results of the experiments conducted using Caltech-256 \cite{griffin2007caltech} and MNIST \cite{mnist} datasets \new{are reported and comparisons with} state-of-the-art outlier detection models \cite{kim2009observe,you2017provable_novelty,xia2015learning_novelty_fig5,sabokrou2018adversarially_alocc, tsakiris2015dual,lerman2015robust,xu2010robust,rahmani2017coherence,liu2010robust} are provided.
%For consistency and comparison purposes, a similar evaluation approach reported in \red{\cite{kim2009observe,you2017provable_novelty,xia2015learning_novelty_fig5,sabokrou2018adversarially_alocc}} is used for the outlier detection performance of our method using Caltech-256\red{\cite{griffin2007caltech}} and MNIST\red{\cite{mnist}} datasets.

\noindent\textbf{Results on Caltech-256.} \blue{Figure~\ref{fig:example_images}b \new{ shows outlier examples reconstructed} using $\mathcal{G}$. It is interesting to observe that although the generated images are of reasonably good quality, our model still depicts superior results in terms of $F_1$ score and area under the curve (AUC), as listed in Table \ref{tab:auc_f1_caltech}}, which demonstrates \bluetwo{that our model is robust to the \textit{over-training} of $\mathcal{G}$.}

%====================================== F1 score MNIST
\begin{figure}[b]
\begin{center}
   \includegraphics[width=0.8\linewidth]{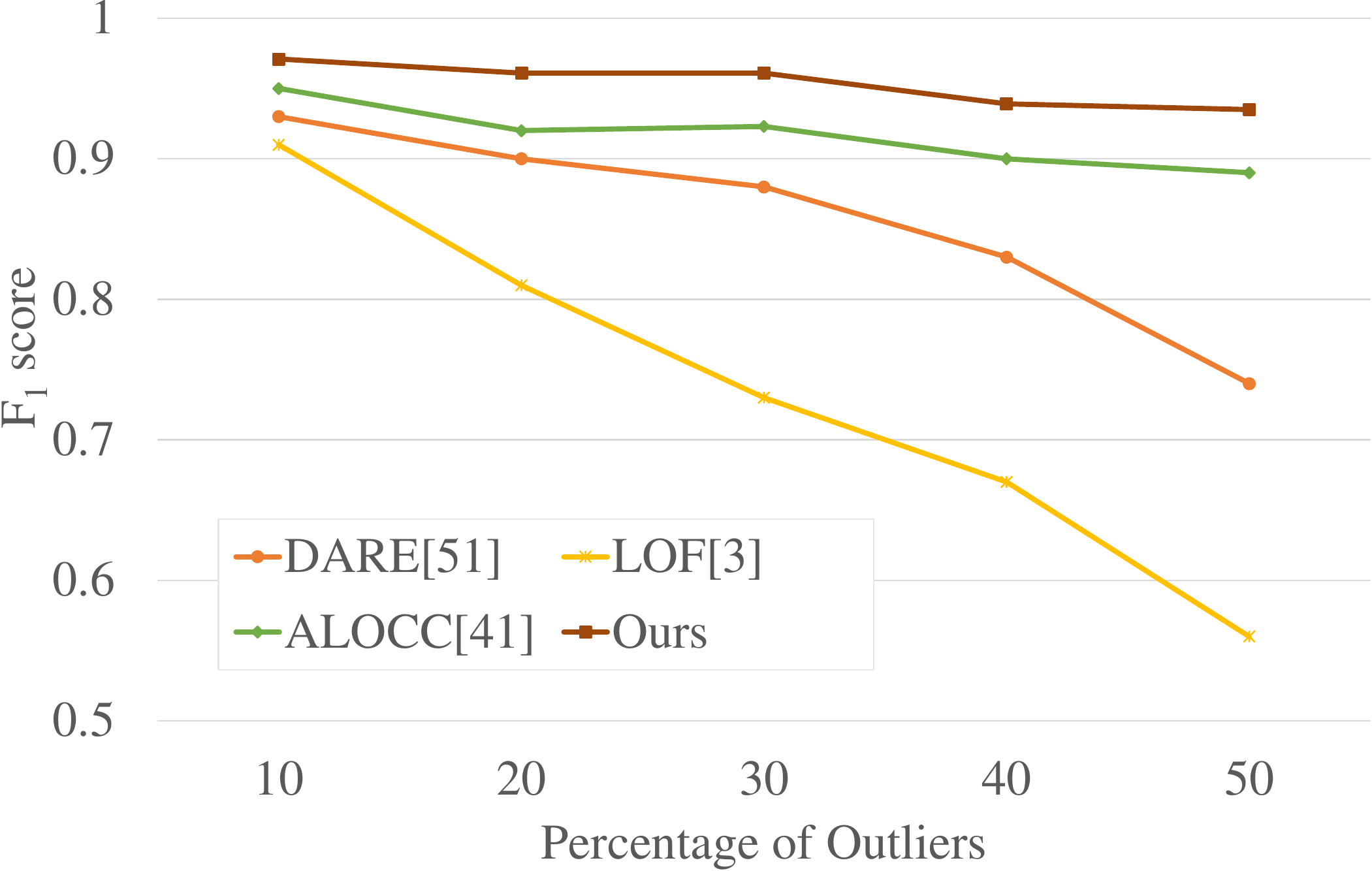}
\end{center}
   \caption{\red{$F_1$ score results on MNIST dataset. Compared to state-of-the-art, our method retains superior performance even with an increased percentage of outliers at test time.}
}
\label{fig:f1_score_mnist}
\end{figure}
%====================================== mnist_score_distribution
\begin{figure}[b]
\begin{center}
%\fbox{\rule{0pt}{2in} \rule{0.8\linewidth}{0pt}}
   \includegraphics[width=0.8\linewidth]{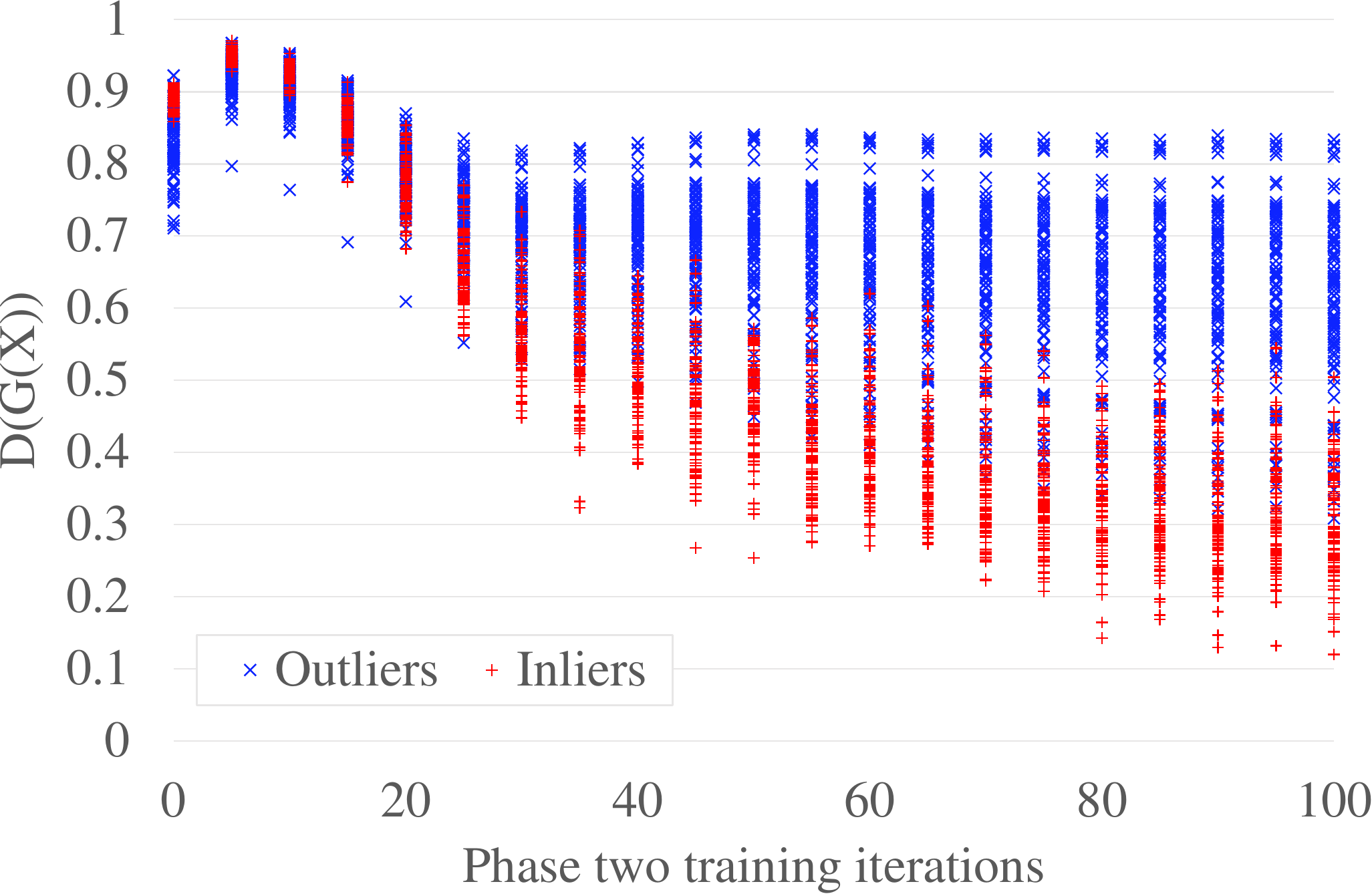}
\end{center}
   \caption{Anomaly score distribution on MNIST dataset over various training iterations of our framework. Divisibility of inliers and outliers is improved significantly as phase two of the training proceeds. 
%   Both inlier and outlier classes are visibly separable in terms of the anomaly scores output by our model. 
}
\label{fig:mnist_score_distribution}
\end{figure}
\noindent\textbf{Results on MNIST.} As it is a well-studied dataset, various outlier detection related works use MNIST as a stepping-stone to evaluate their approaches. Following \cite{xia2015learning_novelty_fig5,breunig2000lof_fig5,sabokrou2018adversarially_alocc}, we also report $F_1$ score as an evaluation metric of our method on this dataset. A comparison provided in \red{Figure~\ref{fig:f1_score_mnist}} shows that our approach performs robustly to detect outliers even when the percentage of outliers is increased. 
%With our proposed framework, the ($\mathcal{D}$) enhances its distinguishability of outliers by noticing even subtle abnormalities in the images generated by the ($\mathcal{G}$), hence yields better results.
An insight of the performance improvement by our approach is shown in Figure~\ref{fig:mnist_score_distribution}. It can be observed that as the phase two training \bluetwo{continues}, score distribution of inliers and outliers output by our network smoothly distributes to a wider range.
%\redtwo{and brings \blue{pushes} the EER threshold \redtwo{to around 0.5} towards the center.}
% \redtwo{Interestingly, we observed this phenomenon in almost every experiment across all three evaluated datasets.}

\subsection{Anomaly Detection in Videos}
One-class classifiers are finding their best applications in the domain of anomaly detection \bluetwo{for surveillance purposes} \cite{sultani2018real_novelty,tudor2017unmasking_novelty,dutta2015online,zhang2016video_novelty,ravanbakhsh2019training,ravanbakhsh2017abnormal_novelty}. 
% \redtwo{Such systems are gaining huge attention due to their massive global scope.}
However, this task is more complicated than the outlier detection because of the involvement of moving objects, which cause variations in appearance. 
%In consistency with the previous works [\red{references}], we report our frame-level anomaly detection results on UCSD Ped2 video anomaly dataset. 

\noindent\textbf{Experimental setup.}
Each frame $I$ of the Ped2 dataset is divided into grayscale patches $X_{I} = \{X_1, X_2, ... , X_n\}$ of size $45 \times 45$ pixels. Normal videos, which only contain scenes of walking pedestrians, are used to extract training patches. Test patches are extracted from abnormal videos which contain abnormal as well as normal scenes. \red{In order to remove unnecessary inference of the patches, a motion detection criteria based on frame difference is set to discard patches without motion.} 
%Once the training of our model is completed, output of the discriminator, evaluated using \red{Equation \ref{eq:frame_anomaly_score}}, is directly taken as the anomaly score for each test patch. 
A maximum of all patch-level anomaly scores is declared as the frame-level anomaly score of that particular frame as:
\begin{equation}\label{eq:frame_anomaly_score}
A_{I} = \underset{X}{\text{max}}\:\mathcal{D}(\mathcal{G}(X)) \text{, where $X \in X_{I}$}
\end{equation}
\noindent\textbf{Performance evaluation.} Frame-level AUC and EER are the two evaluation metrics used to compare our approach with a series of existing works \cite{tudor2017unmasking_novelty,luo2017revisit_novelty,nguyen2019hybrid,hinami2017joint_novelty,liu2018classifier_novelty,ravanbakhsh2017abnormal_novelty,luo2017remembering,Gong_2019_ICCV,ravanbakhsh2018plug_novelty,sun2017online,hasan2016learning_novelty,liu2018future_novelty,xu2015learning_denoise,Nguyen_2019_ICCV,zhang2016video_novelty,ionescu2019object,zhao2017spatio,sabokrou2018adversarially_alocc} published within last \red{5 years}. The corresponding results provided in \red{Table \ref{Tab:EER}} and Table \ref{Tab:AUC} show that our method outperforms recent state-of-the-art methodologies in the task of anomaly detection. \blue{Comparing with the baseline, our approach achieves an absolute gain of 5.2\% in terms of AUC. Examples of the reconstructed patches are provided in Figure~\ref{fig:example_images}. As shown in Figure~\ref{fig:example_images}b, although $\mathcal{G}$ generates \new{noticeably} good reconstructions of anomalous inputs, due to the presence of our proposed pseudo-anomaly module, $\mathcal{D}$ \bluetwo{gets to learn the underlying patterns of reconstructed anomalous images}. This is why, in contrast to the baseline, our framework provides consistent performance across a wide range of training epochs (Figure~\ref{fig:AUC_Comparison}).}
\begin{table}[]
\resizebox{\linewidth}{!}{%
\begin{tabular}{cccc}
RE\cite{sabokrou2016video}       & AbnormalGAN\cite{ravanbakhsh2017abnormal_novelty}  & Ravanbakhsh\cite{ravanbakhsh2019training} & Dan Xu\cite{xu2015learning_denoise}       \\ \hline
15\%     & 13\%         & 14\%        & 17\%         \\ \hline\hline
Sabokrou\cite{sabokrou2015real} & Deep-cascade\cite{sabokrou2017deep_novelty} & ALOCC\cite{sabokrou2018adversarially_alocc}       & Ours         \\ \hline
19\%     & {\ul 9\%}    & 13\%        & \textbf{7\%}
\end{tabular}}
\caption{EER results comparison with existing works on UCSD Ped2 dataset. \new{Lower numbers mean better results.}}
\label{Tab:EER}
\end{table}

\begin{table}[b]
\resizebox{\linewidth}{!}{%
\begin{tabular}{ll|ll}
\multicolumn{1}{l}{Method} & \multicolumn{1}{c|}{AUC} & \multicolumn{1}{l}{Method} & \multicolumn{1}{c}{AUC} \\ \hline
Unmasking\cite{tudor2017unmasking_novelty}                  & 82.2\%                   & TSC\cite{luo2017revisit_novelty}                        & 92.2\%                  \\
HybridDN\cite{nguyen2019hybrid}                   & 84.3\%                   & FRCN action\cite{hinami2017joint_novelty}                & 92.2\%                  \\
Liu et al\cite{liu2018classifier_novelty}                  & 87.5\%                   & AbnormalGAN\cite{ravanbakhsh2017abnormal_novelty}                & 93.5\%                  \\
ConvLSTM\-AE\cite{luo2017remembering}                & 88.1\%                   & MemAE\cite{Gong_2019_ICCV} & 94.1\%                  \\
Ravanbakhsh et al\cite{ravanbakhsh2018plug_novelty}          & 88.4\%                   & GrowingGas\cite{sun2017online}                 & 94.1\%                  \\
ConvAE\cite{hasan2016learning_novelty}                     & 90\%                     & FFP\cite{liu2018future_novelty}      & 95.4\%                  \\
AMDN\cite{xu2015learning_denoise}                       & 90.8\%                    & ConvAE+UNet\cite{Nguyen_2019_ICCV}                & 96.2\%                  \\
Hashing Filters\cite{zhang2016video_novelty}            & 91\%                     &   STAN\cite{lee2018stan}        & 96.5\%     \\
AE\_Conv3D\cite{zhao2017spatio}                 & 91.2\%                   &  Object-centric\cite{ionescu2019object}   &  \underline{97.8\%}                        \\ \hline
Baseline  & 92.9\%  & Ours & \textbf{98.1\%}
\end{tabular}
}
\caption{Frame-level AUC comparison on UCSD Ped2 dataset with state-of-the-art works published in last 5 years. Best and second best performances are highlighted as bold and underlined, respectively.}
\label{Tab:AUC}
\end{table}
\subsection{Discussion}
\begin{figure}[t]
\begin{center}
   \includegraphics[width=0.75\linewidth]{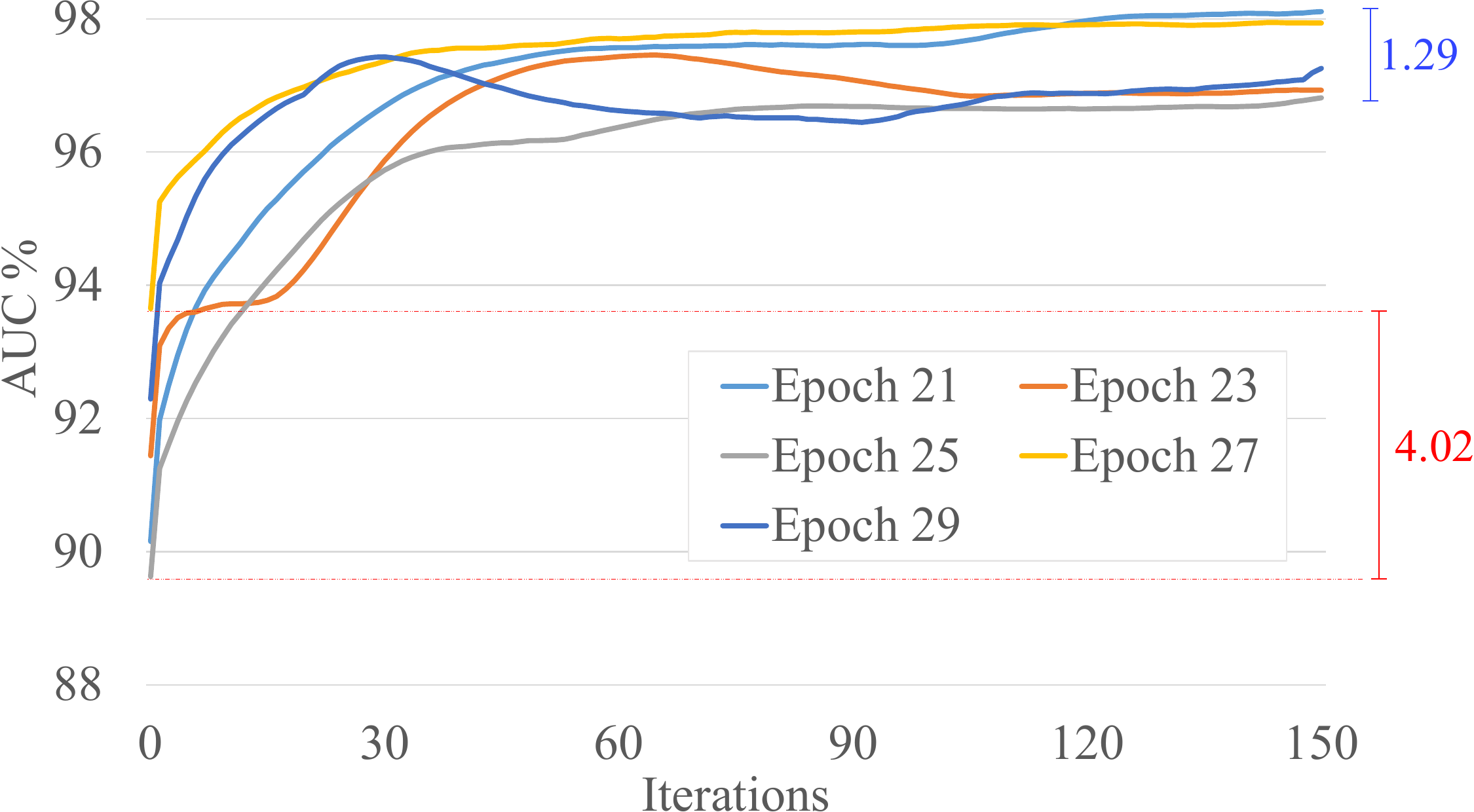}
\end{center}
   \caption{ The plot shows frame level AUC performance of our phase two training, starting after different epochs of phase one (baseline) training. \new{The model after phase two training shows significantly less variance than baseline/phase one.}}
   
%   Compared to baseline, phase two training achieves low variance in convergence.  even if the training in phase one is stopped at arbitrary epochs. Baseline (at iteration=0) shows high fluctuations even between consecutive training epochs}

\label{fig:ucsd_high_epoch_comparison}
\end{figure}

\noindent\textbf{When to stop phase one training?} The convergence of our framework is not strictly dependent on phase one training. \red{Figure~\ref{fig:ucsd_high_epoch_comparison}} shows the AUC performance of phase two training applied after various epochs of phase one on Ped2 dataset \cite{chan2008ucsd}. Values plotted at \red{iterations = 0}, representing the performance of the baseline, show a high variance.  Interestingly, it can be seen that after few iterations into \green{phase two} training of our proposed approach, the model starts to converge better. Irrespective of the initial epoch in phase one training, models converged successfully showing consistent AUC performances.

\begin{figure}[t]
\begin{center}
     \subfloat[\new{Experiments with the $\mathcal{G}^{old}$ taken across first 10 epochs, where $\mathcal{G}$ is kept fixed. }]{%
       \includegraphics[width=0.75\linewidth]{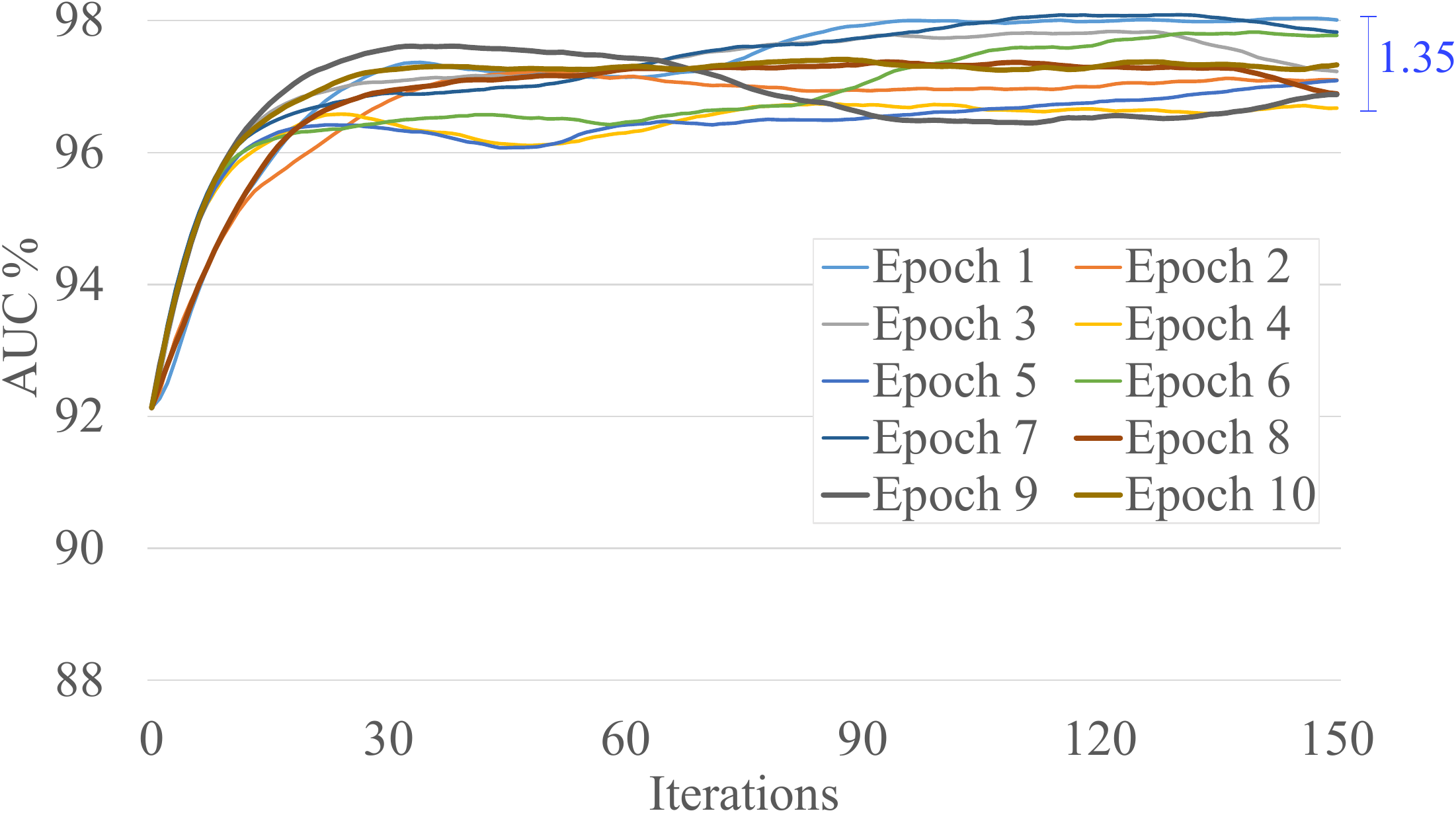}
       \label{fig:ucsd_low_epoch_comparison}
     }
     \hfill
     \subfloat[\new{Experiments with the $\mathcal{G}^{old}$ obtained at various arbitrary epochs by averaging the parameters of generators from all previous epochs.}]
     {%
       \includegraphics[width=0.75\linewidth]{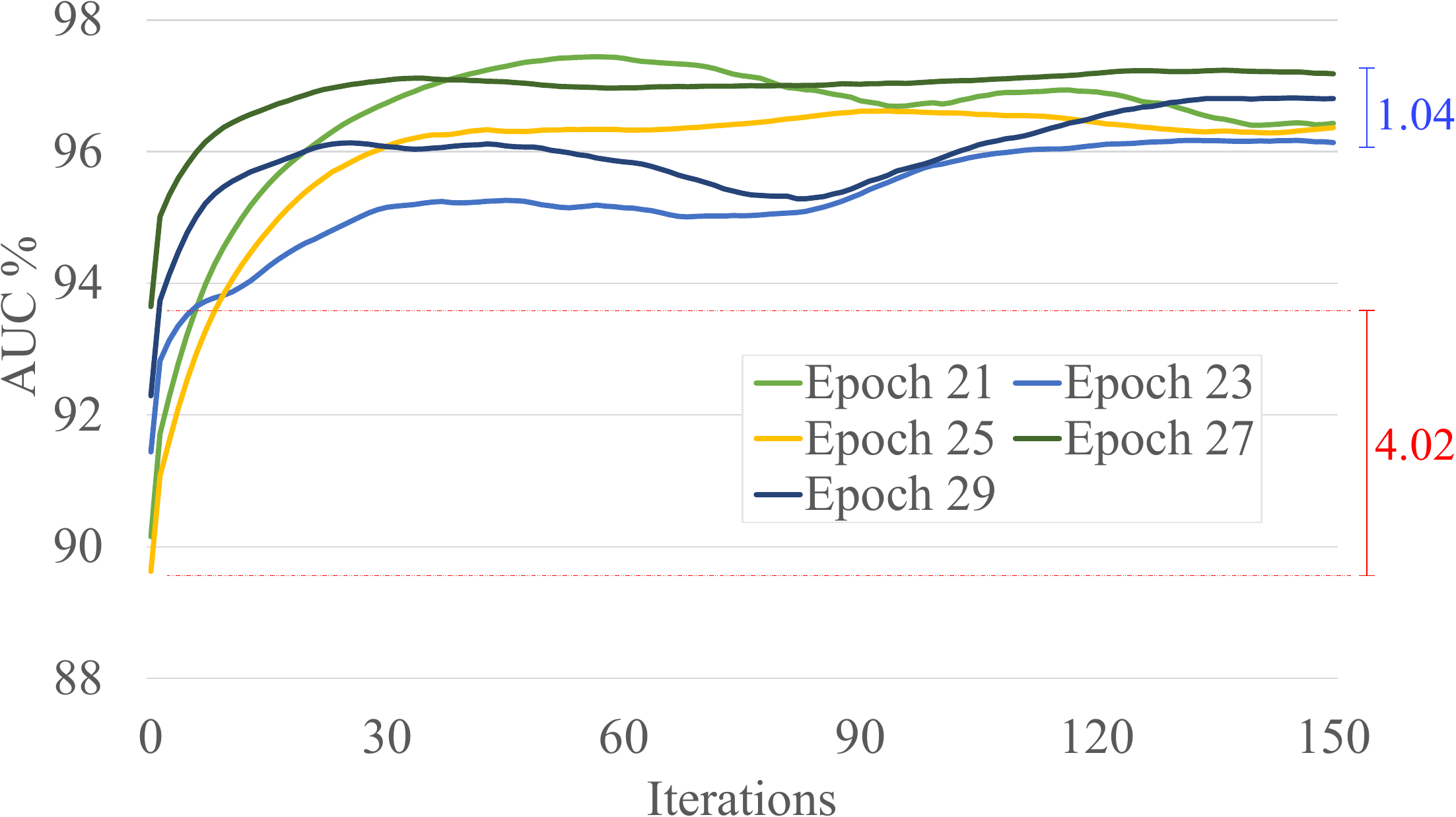}
       \label{fig:ucsd_average_low_epoch_comparison}
     }
\end{center}
     \caption{Results from a series of experiments on UCSD Ped2 dataset show that our framework is not dependent on a strict choice of epoch number for  $\mathcal{G}^{old}$. In (a), various $\mathcal{G}^{old}$ selected at a varied range of epochs are experimented with a fixed $\mathcal{G}$. In (b), an average of parameters from all past generators is taken as $\mathcal{G}^{old}$.}
     \label{fig:low_epoch_comparison}
   \end{figure}

\noindent\textbf{When to stop phase two training?} As seen in \red{Figure~\ref{fig:ucsd_high_epoch_comparison}} and Figure~\ref{fig:low_epoch_comparison}, it can be observed that once a specific model is converged, further iterations do not deteriorate its performance. Hence, a model can be trained for any number of iterations as deemed necessary.
%====================================== ucsd_high_epoch_comparison

\noindent\textbf{Which low epoch generator is better?} For the selection of $\mathcal{G}^{old}$, as mentioned earlier, the generator after the $1^{st}$ epoch of training was arbitrarily chosen in our experiments. This selection is \new{ intuitive and mostly based on the fact that the generator has seen all dataset once. In addition, we visually observed that after first epoch, although the generator was capable of reconstructing its input, the quality was not ‘good enough’, which is a suitable property for $\mathcal{G}^{old}$ in our model.} However, \new{this way of selection} is not a generalized solution across various datasets. 
% Some datasets have lesser training instances than the others and hence a model may not learn enough to produce results other than noise after being trained for just one epoch.
Hence, to investigate the matter further, we evaluate a range of low epoch numbers as candidates for $\mathcal{G}^{old}$. \red{The baseline epoch of $\mathcal{G}$} is kept fixed throughout this experiment. Results in Figure~\ref{fig:ucsd_low_epoch_comparison} show that irrespective of the low epoch number chosen as $\mathcal{G}^{old}$, the model converges and achieves state-of-the-art or comparable AUC. In pursuit of another more systematic way to obtain $\mathcal{G}^{old}$, we also explored the possibility of using average parameters of all previous $\mathcal{G}$ models. Hence, for each given epoch of the baseline that we pick as $\mathcal{G}$, \new{a} $\mathcal{G}^{old}$ is formulated by taking an average of all previous $\mathcal{G}$ models until that point. The results plotted in Figure~\ref{fig:ucsd_average_low_epoch_comparison} \new{show that such $\mathcal{G}^{old}$} also depicts comparable performances. 
% \redtwo{All variants of the models achieved at least 96\% AUC performance on UCSD Ped2 dataset.}
\new{Note that} this formulation completely eradicates the need of handpicking a specific epoch number for $\mathcal{G}^{old}$, thus making our formulation generic towards the size of a training dataset. 

%====================================== ucsd_low_epoch_comparison

% \begin{figure}[]
% \begin{center}
%   \includegraphics[width=0.75\linewidth]{ucsd_low_epoch_comparison.png}
% \end{center}
%   \caption{\red{ The plot shows frame level AUC performance of our Phase two training, starting after different epochs of phase one (baseline) training. Every training in phase two converged within a range of 96.8\% - 98.1\% AUC.
   
%   Choice of Gold does not put any significant impact on the performance}
% }
% \label{fig:ucsd_low_epoch_comparison}
% \end{figure}
% %====================================== ucsd_average_epoch_comparison
% \begin{figure}
% \begin{center}
%   \includegraphics[width=0.75\linewidth]{ucsd_average_low_epoch_comparison.png}
   
% \end{center}
%   \caption{\red{A generic approach to choose Gold by averaging the weighs and biases of generator until the most recent state. Model at each training epoch successfully converged to achieved more than 00\% AUC}
% }
% \label{fig:ucsd_average_low_epoch_comparison}
% \end{figure}
\subsection{Ablation}
Ablation results of our framework on UCSD Ped2 dataset \cite{chan2008ucsd} are summarized in \red{Table \ref{tab:ablation}}. As shown, while each input component of our training model (\red{i.e. real images $X$, high quality reconstructions $\hat{X}$, low quality reconstructions $\hat{X}^{low}$, and pseudo anomaly reconstructions $\hat{X}^{pseudo}$)} contributes towards a robust training, removing any of these at a time still shows better performance than the baseline. One interesting observation can be seen in the \new{fourth column of the phase two} training results. In this case, the performance is measured after we remove the last step of pseudo-anomaly module, which is responsible \new{for providing} regenerated pseudo-anomaly ($\hat{X}^{pseudo}$) through $\mathcal{G}$, as in Equation \ref{eq:pseudo_reconstruction}. Hence, by removing this part, the fake anomalies ($\hat{\bar{X}}$) obtained using Equation \ref{eq:fake_anomaly} are channeled directly to the discriminator as one of the two sets of bad reconstruction examples. With this configuration, the performance deteriorates significantly (i.e. 9.6\% drop in the AUC). The model shows even worse performance than the baseline after \green{phase one} training. This shows the significance of our proposed pseudo-anomaly module. Once pseudo-anomalies are created within the module, it is necessary to obtain a regeneration result of these by inferring $\mathcal{G}$.  
% \redtwo{ This way the $\mathcal{D}$ learns about the behavior of $\mathcal{G}$ in the case of an unusual input and hence complements it during the test time, resulting in a stable anomaly detection model.} 
\bluetwo{This helps $\mathcal{D}$ to learn the underlying patterns of reconstructed anomalous images, which results in a \new{more robust} anomaly detection model.}
\begin{table}[]
\resizebox{\linewidth}{!}{
\begin{tabular}{c|c|ccccc}
                                      & \green{Phase one} & \multicolumn{5}{c}{\green{Phase two} }               \\ \hline
$X$                                   & \ding{51}                & -      & \ding{51}     & \ding{51}     & \ding{51}     & \ding{51}              \\
$\hat{X}$                             & \ding{51}                & \ding{51}     & \ding{51}     & \ding{51}     & \ding{51}     & \ding{51}              \\
$\hat{X}^{low}$                       & -                & \ding{51}     & \ding{51}     & -      & \ding{51}     & \ding{51}              \\
$\hat{X}^{pseudo}$                    & -                & -      & -      & \ding{51}     & -      & \ding{51}              \\
$\hat{\bar{X}}$ as $\hat{X}^{pseudo}$ & -                & -      & -      & -      & \ding{51}     & -               \\ \hline
AUC                                   & 92.9\%           & 94.4\% & 95.1\% & 95.9\% & 88.5\% & \textbf{98.1\%}
\end{tabular}}
\caption{Frame-level AUC performance ablation of our framework on UCSD Ped2 dataset. 
% The importance of each training component is shown. Removal of the pseudo-anomaly regeneration step, (Equation \ref{eq:pseudo_reconstruction}), from pseudo-anomaly module shows noticeable performance degradation
}
\label{tab:ablation}
\end{table}
\section{Conclusion}
This paper presents an \new{adversarially learned} approach in which both the generator ($\mathcal{G}$) and the discriminator ($\mathcal{D}$) are utilized to perform a stable and robust anomaly detection. A unified $\mathcal{G}$ and $\mathcal{D}$ model employed towards such problems often produces unstable results due to the adversary. However, we attempted to tweak the basic role of the discriminator from distinguishing between real and fake to discriminating between good and bad quality reconstructions, a formulation that aligns \bluetwo{well} with the philosophy of conventional anomaly detection using generative networks. We also propose a pseudo-anomaly module which is employed to create fake anomaly examples from normal training data. These fake anomaly examples help $\mathcal{D}$ to learn about the behavior of $\mathcal{G}$ \new{in the case of} unusual input data.

Our extensive experimentation shows that the approach not only generates stable results across a wide range of training epochs but also outperforms a series of state-of-the-art methods \cite{tudor2017unmasking_novelty,luo2017revisit_novelty,nguyen2019hybrid,hinami2017joint_novelty,liu2018classifier_novelty,ravanbakhsh2017abnormal_novelty,luo2017remembering,Gong_2019_ICCV,ravanbakhsh2018plug_novelty,sun2017online,hasan2016learning_novelty,liu2018future_novelty,xu2015learning_denoise,Nguyen_2019_ICCV,zhang2016video_novelty,ionescu2019object,zhao2017spatio,sabokrou2018adversarially_alocc} for outliers and anomaly detection. 
% \redtwo{We aim to explore the possibilities of using our idea of making $\mathcal{D}$ and $\mathcal{G}$ complement each other for a more robust solution towards various other domains such as object detection and tracking.}

\section{Acknowledgment}
This work was supported by the ICT R\&D program of MSIP/IITP. [2017-0-00306, Development of Multimodal Sensor-based Intelligent Systems for Outdoor Surveillance Robots]. \new{Also, we thank HoChul Shin, Ki-In Na, Hamza Saleem, Ayesha Zaheer, Arif Mahmood, and Shah Nawaz for the discussions and support in improving our work.}

{\small
\bibliographystyle{ieee_fullname}
\bibliography{egpaper_final}

\begin{thebibliography}{10}\itemsep=-1pt

\bibitem{basharat2008learning_realworld7}
Arslan Basharat, Alexei Gritai, and Mubarak Shah.
\newblock Learning object motion patterns for anomaly detection and improved
  object detection.
\newblock In {\em 2008 IEEE Conference on Computer Vision and Pattern
  Recognition}, pages 1--8. IEEE, 2008.

\bibitem{bergadano2019keyed}
Francesco Bergadano.
\newblock Keyed learning: An adversarial learning framework—formalization,
  challenges, and anomaly detection applications.
\newblock {\em ETRI Journal}, 41(5):608--618, 2019.

\bibitem{breunig2000lof_fig5}
Markus~M Breunig, Hans-Peter Kriegel, Raymond~T Ng, and J{\"o}rg Sander.
\newblock Lof: identifying density-based local outliers.
\newblock In {\em ACM sigmod record}, volume~29, pages 93--104. ACM, 2000.

\bibitem{chan2008ucsd}
Antoni Chan and Nuno Vasconcelos.
\newblock Ucsd pedestrian dataset.
\newblock {\em IEEE Trans. on Pattern Analysis and Machine Intelligence
  (TPAMI)}, 30(5):909--926, 2008.

\bibitem{cui2011abnormal_realworld10}
Xinyi Cui, Qingshan Liu, Mingchen Gao, and Dimitris~N Metaxas.
\newblock Abnormal detection using interaction energy potentials.
\newblock In {\em CVPR 2011}, pages 3161--3167. IEEE, 2011.

\bibitem{dutta2015online}
Jayanta~Kumar Dutta and Bonny Banerjee.
\newblock Online detection of abnormal events using incremental coding length.
\newblock In {\em Twenty-Ninth AAAI Conference on Artificial Intelligence},
  2015.

\bibitem{Gong_2019_ICCV}
Dong Gong, Lingqiao Liu, Vuong Le, Budhaditya Saha, Moussa~Reda Mansour, Svetha
  Venkatesh, and Anton van~den Hengel.
\newblock Memorizing normality to detect anomaly: Memory-augmented deep
  autoencoder for unsupervised anomaly detection.
\newblock In {\em The IEEE International Conference on Computer Vision (ICCV)},
  October 2019.

\bibitem{goodfellow2014generative_gan}
Ian Goodfellow, Jean Pouget-Abadie, Mehdi Mirza, Bing Xu, David Warde-Farley,
  Sherjil Ozair, Aaron Courville, and Yoshua Bengio.
\newblock Generative adversarial nets.
\newblock In {\em Advances in neural information processing systems}, pages
  2672--2680, 2014.

\bibitem{griffin2007caltech}
Gregory Griffin, Alex Holub, and Pietro Perona.
\newblock Caltech-256 object category dataset.
\newblock 2007.

\bibitem{hasan2016learning_novelty}
Mahmudul Hasan, Jonghyun Choi, Jan Neumann, Amit~K Roy-Chowdhury, and Larry~S
  Davis.
\newblock Learning temporal regularity in video sequences.
\newblock In {\em Proceedings of the IEEE conference on computer vision and
  pattern recognition}, pages 733--742, 2016.

\bibitem{hinami2017joint_novelty}
Ryota Hinami, Tao Mei, and Shin'ichi Satoh.
\newblock Joint detection and recounting of abnormal events by learning deep
  generic knowledge.
\newblock In {\em Proceedings of the IEEE International Conference on Computer
  Vision}, pages 3619--3627, 2017.

\bibitem{hou2017tube_realworld20}
Rui Hou, Chen Chen, and Mubarak Shah.
\newblock Tube convolutional neural network (t-cnn) for action detection in
  videos.
\newblock In {\em Proceedings of the IEEE International Conference on Computer
  Vision}, pages 5822--5831, 2017.

\bibitem{ionescu2019object}
Radu~Tudor Ionescu, Fahad~Shahbaz Khan, Mariana-Iuliana Georgescu, and Ling
  Shao.
\newblock Object-centric auto-encoders and dummy anomalies for abnormal event
  detection in video.
\newblock In {\em Proceedings of the IEEE Conference on Computer Vision and
  Pattern Recognition}, pages 7842--7851, 2019.

\bibitem{kim2009observe}
Jaechul Kim and Kristen Grauman.
\newblock Observe locally, infer globally: a space-time mrf for detecting
  abnormal activities with incremental updates.
\newblock In {\em 2009 IEEE Conference on Computer Vision and Pattern
  Recognition}, pages 2921--2928. IEEE, 2009.

\bibitem{kingma2014adam}
Diederik~P Kingma and Jimmy Ba.
\newblock Adam: A method for stochastic optimization.
\newblock {\em arXiv preprint arXiv:1412.6980}, 2014.

\bibitem{kratz2009anomaly_realworld26}
Louis Kratz and Ko Nishino.
\newblock Anomaly detection in extremely crowded scenes using spatio-temporal
  motion pattern models.
\newblock In {\em 2009 IEEE Conference on Computer Vision and Pattern
  Recognition}, pages 1446--1453. IEEE, 2009.

\bibitem{lawson2017finding}
Wallace Lawson, Esube Bekele, and Keith Sullivan.
\newblock Finding anomalies with generative adversarial networks for a
  patrolbot.
\newblock In {\em Proceedings of the IEEE Conference on Computer Vision and
  Pattern Recognition Workshops}, pages 12--13, 2017.

\bibitem{mnist}
Yann LeCun, Corinna Cortes, and Christopher J.C.~Burges.
\newblock Mnist handwritten digit database.
\newblock {\em AT\&T Labs [Online].Available: http://yann. lecun.
  com/exdb/mnist}, 2010.

\bibitem{lee2018stan}
Sangmin Lee, Hak~Gu Kim, and Yong~Man Ro.
\newblock Stan: Spatio-temporal adversarial networks for abnormal event
  detection.
\newblock In {\em 2018 IEEE International Conference on Acoustics, Speech and
  Signal Processing (ICASSP)}, pages 1323--1327. IEEE, 2018.

\bibitem{lerman2015robust}
Gilad Lerman, Michael~B McCoy, Joel~A Tropp, and Teng Zhang.
\newblock Robust computation of linear models by convex relaxation.
\newblock {\em Foundations of Computational Mathematics}, 15(2):363--410, 2015.

\bibitem{liu2010robust}
Guangcan Liu, Zhouchen Lin, and Yong Yu.
\newblock Robust subspace segmentation by low-rank representation.
\newblock In {\em ICML}, volume~1, page~8, 2010.

\bibitem{liu2018future_novelty}
Wen Liu, Weixin Luo, Dongze Lian, and Shenghua Gao.
\newblock Future frame prediction for anomaly detection--a new baseline.
\newblock In {\em Proceedings of the IEEE Conference on Computer Vision and
  Pattern Recognition}, pages 6536--6545, 2018.

\bibitem{liu2018classifier_novelty}
Yusha Liu, Chun-Liang Li, and Barnab{\'a}s P{\'o}czos.
\newblock Classifier two sample test for video anomaly detections.
\newblock In {\em BMVC}, page~71, 2018.

\bibitem{luo2017remembering}
Weixin Luo, Wen Liu, and Shenghua Gao.
\newblock Remembering history with convolutional lstm for anomaly detection.
\newblock In {\em 2017 IEEE International Conference on Multimedia and Expo
  (ICME)}, pages 439--444. IEEE, 2017.

\bibitem{luo2017revisit_novelty}
Weixin Luo, Wen Liu, and Shenghua Gao.
\newblock A revisit of sparse coding based anomaly detection in stacked rnn
  framework.
\newblock In {\em Proceedings of the IEEE International Conference on Computer
  Vision}, pages 341--349, 2017.

\bibitem{medioni2001event_twostream34}
G{\'e}rard Medioni, Isaac Cohen, Fran{\c{c}}ois Br{\'e}mond, Somboon Hongeng,
  and Ramakant Nevatia.
\newblock Event detection and analysis from video streams.
\newblock {\em IEEE Transactions on pattern analysis and machine intelligence},
  23(8):873--889, 2001.

\bibitem{Nguyen_2019_ICCV}
Trong-Nguyen Nguyen and Jean Meunier.
\newblock Anomaly detection in video sequence with appearance-motion
  correspondence.
\newblock In {\em The IEEE International Conference on Computer Vision (ICCV)},
  October 2019.

\bibitem{nguyen2019hybrid}
Trong~Nguyen Nguyen and Jean Meunier.
\newblock Hybrid deep network for anomaly detection.
\newblock {\em arXiv preprint arXiv:1908.06347}, 2019.

\bibitem{paszke2017automatic}
Adam Paszke, Sam Gross, Soumith Chintala, Gregory Chanan, Edward Yang, Zachary
  DeVito, Zeming Lin, Alban Desmaison, Luca Antiga, and Adam Lerer.
\newblock Automatic differentiation in pytorch.
\newblock 2017.

\bibitem{pathak2016context_adversarial_good}
Deepak Pathak, Philipp Krahenbuhl, Jeff Donahue, Trevor Darrell, and Alexei~A
  Efros.
\newblock Context encoders: Feature learning by inpainting.
\newblock In {\em Proceedings of the IEEE conference on computer vision and
  pattern recognition}, pages 2536--2544, 2016.

\bibitem{piciarelli2008trajectory_twostream36}
Claudio Piciarelli, Christian Micheloni, and Gian~Luca Foresti.
\newblock Trajectory-based anomalous event detection.
\newblock {\em IEEE Transactions on Circuits and Systems for video Technology},
  18(11):1544--1554, 2008.

\bibitem{radford2015unsupervised_good_adversarial}
Alec Radford, Luke Metz, and Soumith Chintala.
\newblock Unsupervised representation learning with deep convolutional
  generative adversarial networks.
\newblock {\em arXiv preprint arXiv:1511.06434}, 2015.

\bibitem{rahmani2017coherence}
Mostafa Rahmani and George~K Atia.
\newblock Coherence pursuit: Fast, simple, and robust principal component
  analysis.
\newblock {\em IEEE Transactions on Signal Processing}, 65(23):6260--6275,
  2017.

\bibitem{ravanbakhsh2018plug_novelty}
Mahdyar Ravanbakhsh, Moin Nabi, Hossein Mousavi, Enver Sangineto, and Nicu
  Sebe.
\newblock Plug-and-play cnn for crowd motion analysis: An application in
  abnormal event detection.
\newblock In {\em 2018 IEEE Winter Conference on Applications of Computer
  Vision (WACV)}, pages 1689--1698. IEEE, 2018.

\bibitem{ravanbakhsh2017abnormal_novelty}
Mahdyar Ravanbakhsh, Moin Nabi, Enver Sangineto, Lucio Marcenaro, Carlo
  Regazzoni, and Nicu Sebe.
\newblock Abnormal event detection in videos using generative adversarial nets.
\newblock In {\em 2017 IEEE International Conference on Image Processing
  (ICIP)}, pages 1577--1581. IEEE, 2017.

\bibitem{ravanbakhsh2019training}
Mahdyar Ravanbakhsh, Enver Sangineto, Moin Nabi, and Nicu Sebe.
\newblock Training adversarial discriminators for cross-channel abnormal event
  detection in crowds.
\newblock In {\em 2019 IEEE Winter Conference on Applications of Computer
  Vision (WACV)}, pages 1896--1904. IEEE, 2019.

\bibitem{ren2015unsupervised}
Huamin Ren, Weifeng Liu, S{\o}ren~Ingvor Olsen, Sergio Escalera, and Thomas~B
  Moeslund.
\newblock Unsupervised behavior-specific dictionary learning for abnormal event
  detection.
\newblock In {\em BMVC}, pages 28--1, 2015.

\bibitem{sabokrou2016video}
Mohammad Sabokrou, Mahmood Fathy, and Mojtaba Hoseini.
\newblock Video anomaly detection and localisation based on the sparsity and
  reconstruction error of auto-encoder.
\newblock {\em Electronics Letters}, 52(13):1122--1124, 2016.

\bibitem{sabokrou2015real}
Mohammad Sabokrou, Mahmood Fathy, Mojtaba Hoseini, and Reinhard Klette.
\newblock Real-time anomaly detection and localization in crowded scenes.
\newblock In {\em Proceedings of the IEEE conference on computer vision and
  pattern recognition workshops}, pages 56--62, 2015.

\bibitem{sabokrou2017deep_novelty}
Mohammad Sabokrou, Mohsen Fayyaz, Mahmood Fathy, and Reinhard Klette.
\newblock Deep-cascade: Cascading 3d deep neural networks for fast anomaly
  detection and localization in crowded scenes.
\newblock {\em IEEE Transactions on Image Processing}, 26(4):1992--2004, 2017.

\bibitem{sabokrou2018adversarially_alocc}
Mohammad Sabokrou, Mohammad Khalooei, Mahmood Fathy, and Ehsan Adeli.
\newblock Adversarially learned one-class classifier for novelty detection.
\newblock In {\em Proceedings of the IEEE Conference on Computer Vision and
  Pattern Recognition}, pages 3379--3388, 2018.

\bibitem{schlegl2017unsupervised}
Thomas Schlegl, Philipp Seeb{\"o}ck, Sebastian~M Waldstein, Ursula
  Schmidt-Erfurth, and Georg Langs.
\newblock Unsupervised anomaly detection with generative adversarial networks
  to guide marker discovery.
\newblock In {\em International Conference on Information Processing in Medical
  Imaging}, pages 146--157. Springer, 2017.

\bibitem{Shama_2019_ICCV_good}
Firas Shama, Roey Mechrez, Alon Shoshan, and Lihi Zelnik-Manor.
\newblock Adversarial feedback loop.
\newblock In {\em The IEEE International Conference on Computer Vision (ICCV)},
  October 2019.

\bibitem{smeureanu2017deep_novelty}
Sorina Smeureanu, Radu~Tudor Ionescu, Marius Popescu, and Bogdan Alexe.
\newblock Deep appearance features for abnormal behavior detection in video.
\newblock In {\em International Conference on Image Analysis and Processing},
  pages 779--789. Springer, 2017.

\bibitem{sultani2018real_novelty}
Waqas Sultani, Chen Chen, and Mubarak Shah.
\newblock Real-world anomaly detection in surveillance videos.
\newblock In {\em Proceedings of the IEEE Conference on Computer Vision and
  Pattern Recognition}, pages 6479--6488, 2018.

\bibitem{sun2017online}
Qianru Sun, Hong Liu, and Tatsuya Harada.
\newblock Online growing neural gas for anomaly detection in changing
  surveillance scenes.
\newblock {\em Pattern Recognition}, 64:187--201, 2017.

\bibitem{tsakiris2015dual}
Manolis~C Tsakiris and Ren{\'e} Vidal.
\newblock Dual principal component pursuit.
\newblock In {\em Proceedings of the IEEE International Conference on Computer
  Vision Workshops}, pages 10--18, 2015.

\bibitem{tudor2017unmasking_novelty}
Radu Tudor~Ionescu, Sorina Smeureanu, Bogdan Alexe, and Marius Popescu.
\newblock Unmasking the abnormal events in video.
\newblock In {\em Proceedings of the IEEE International Conference on Computer
  Vision}, pages 2895--2903, 2017.

\bibitem{vincent2008extracting_denoise}
Pascal Vincent, Hugo Larochelle, Yoshua Bengio, and Pierre-Antoine Manzagol.
\newblock Extracting and composing robust features with denoising autoencoders.
\newblock In {\em Proceedings of the 25th international conference on Machine
  learning}, pages 1096--1103. ACM, 2008.

\bibitem{wang2014learning_realworld38}
Jiang Wang, Yang Song, Thomas Leung, Chuck Rosenberg, Jingbin Wang, James
  Philbin, Bo Chen, and Ying Wu.
\newblock Learning fine-grained image similarity with deep ranking.
\newblock In {\em Proceedings of the IEEE Conference on Computer Vision and
  Pattern Recognition}, pages 1386--1393, 2014.

\bibitem{xia2015learning_novelty_fig5}
Yan Xia, Xudong Cao, Fang Wen, Gang Hua, and Jian Sun.
\newblock Learning discriminative reconstructions for unsupervised outlier
  removal.
\newblock In {\em Proceedings of the IEEE International Conference on Computer
  Vision}, pages 1511--1519, 2015.

\bibitem{xu2015learning_denoise}
Dan Xu, Elisa Ricci, Yan Yan, Jingkuan Song, and Nicu Sebe.
\newblock Learning deep representations of appearance and motion for anomalous
  event detection.
\newblock {\em arXiv preprint arXiv:1510.01553}, 2015.

\bibitem{xu2017detecting_denoise}
Dan Xu, Yan Yan, Elisa Ricci, and Nicu Sebe.
\newblock Detecting anomalous events in videos by learning deep representations
  of appearance and motion.
\newblock {\em Computer Vision and Image Understanding}, 156:117--127, 2017.

\bibitem{xu2010robust}
Huan Xu, Constantine Caramanis, and Sujay Sanghavi.
\newblock Robust pca via outlier pursuit.
\newblock In {\em Advances in Neural Information Processing Systems}, pages
  2496--2504, 2010.

\bibitem{you2017provable_novelty}
Chong You, Daniel~P Robinson, and Ren{\'e} Vidal.
\newblock Provable self-representation based outlier detection in a union of
  subspaces.
\newblock In {\em Proceedings of the IEEE Conference on Computer Vision and
  Pattern Recognition}, pages 3395--3404, 2017.

\bibitem{zaheer2018ensemble}
Muhammad~Zaigham Zaheer, Marcella Astrid, Seung-Ik Lee, and Ho~Chul Shin.
\newblock Ensemble grid formation to detect potential anomalous regions using
  context encoders.
\newblock In {\em 2018 18th International Conference on Control, Automation and
  Systems (ICCAS)}, pages 661--665. IEEE, 2018.

\bibitem{zhang2009learning_twostream53}
Tianzhu Zhang, Hanqing Lu, and Stan~Z Li.
\newblock Learning semantic scene models by object classification and
  trajectory clustering.
\newblock In {\em 2009 IEEE Conference on Computer Vision and Pattern
  Recognition}, pages 1940--1947. IEEE, 2009.

\bibitem{zhang2016video_novelty}
Ying Zhang, Huchuan Lu, Lihe Zhang, Xiang Ruan, and Shun Sakai.
\newblock Video anomaly detection based on locality sensitive hashing filters.
\newblock {\em Pattern Recognition}, 59:302--311, 2016.

\bibitem{zhao2017spatio}
Yiru Zhao, Bing Deng, Chen Shen, Yao Liu, Hongtao Lu, and Xian-Sheng Hua.
\newblock Spatio-temporal autoencoder for video anomaly detection.
\newblock In {\em Proceedings of the 25th ACM international conference on
  Multimedia}, pages 1933--1941. ACM, 2017.

\end{thebibliography}
}

\end{document}